\documentclass[10pt,twocolumn,letterpaper]{article}

\usepackage[pagenumbers]{cvpr}              %

\definecolor{cvprblue}{rgb}{0.21,0.49,0.74}
\usepackage[pagebackref,breaklinks,colorlinks,allcolors=cvprblue]{hyperref}

\def\paperID{3808} %
\def\confName{CVPR}
\def\confYear{2025}

\usepackage[accsupp]{axessibility}  %

\usepackage{amsmath}
\usepackage{url}
\usepackage{amssymb}
\usepackage{xcolor}
\usepackage{xspace}
\usepackage{graphicx}
\usepackage{pifont}%
\usepackage{enumitem}

\usepackage{algorithm}
\usepackage[noend]{algpseudocode}

\usepackage{graphicx}
\usepackage{makecell}
\usepackage{booktabs}
\usepackage{tabularx}
\usepackage{multirow}

\usepackage{mathtools} %
\usepackage{rotating}

\def\hide#1{}

\DeclareMathOperator*{\argmax}{argmax}

\newcommand{\escher}{\textsc{Escher}\xspace}
\newcommand{\name}{\textsc{Escher}\xspace}

\newcommand{\Data}{\mathcal{D}}
\newcommand{\Con}{\mathcal{C}}
\newcommand{\score}{\texttt{score}_{\texttt{VLM}}}
\newcommand{\Model}{\theta_{\text{VLM}}}
\newcommand{\Adapter}{w_{\mathcal{Y}}}

\usepackage{adjustbox}

\title{Self-Evolving Visual Concept Library using Vision-Language Critics}

\author{
Atharva Sehgal$^{1}$\thanks{Correspondence to atharvas@utexas.edu. Artifacts available at \url{https://trishullab.github.io/escher-web}}
~~
Patrick Yuan$^{2}$
~~
Ziniu Hu$^{3}$
~~
Yisong Yue$^{3}$
~~
Jennifer J. Sun$^{2}$
~~
Swarat Chaudhuri$^{1}$
\\\\
$^{1}$University of Texas at Austin
~~
$^{2}$Cornell University
~~
$^{3}$California Institute of Technology
}

\begin{document}
\maketitle

\begin{abstract}

We study the problem of building a visual concept library for visual recognition. Building effective visual concept libraries is challenging, as manual definition is labor-intensive, while relying solely on LLMs for concept generation can result in concepts that lack discriminative power or fail to account for the complex interactions between them. Our approach, \escher, takes a library learning perspective to iteratively discover and improve visual concepts. \escher uses a vision-language model (VLM) as a critic to iteratively refine the concept library, including accounting for interactions between concepts and how they affect downstream classifiers. By leveraging the in-context learning abilities of LLMs and the history of performance using various concepts, \escher dynamically improves its concept generation strategy based on the VLM critic's feedback. Finally, \escher does not require any human annotations, and is thus an automated plug-and-play framework. We empirically demonstrate the ability of \escher to learn a concept library for zero-shot, few-shot, and fine-tuning visual classification tasks. This work represents, to our knowledge, the first application of concept library learning to real-world visual tasks.

\end{abstract}

\section{Introduction}\label{sec:introduction}

\begin{figure}
    \centering
    \vspace{0.1in}
    \includegraphics[width=\linewidth]{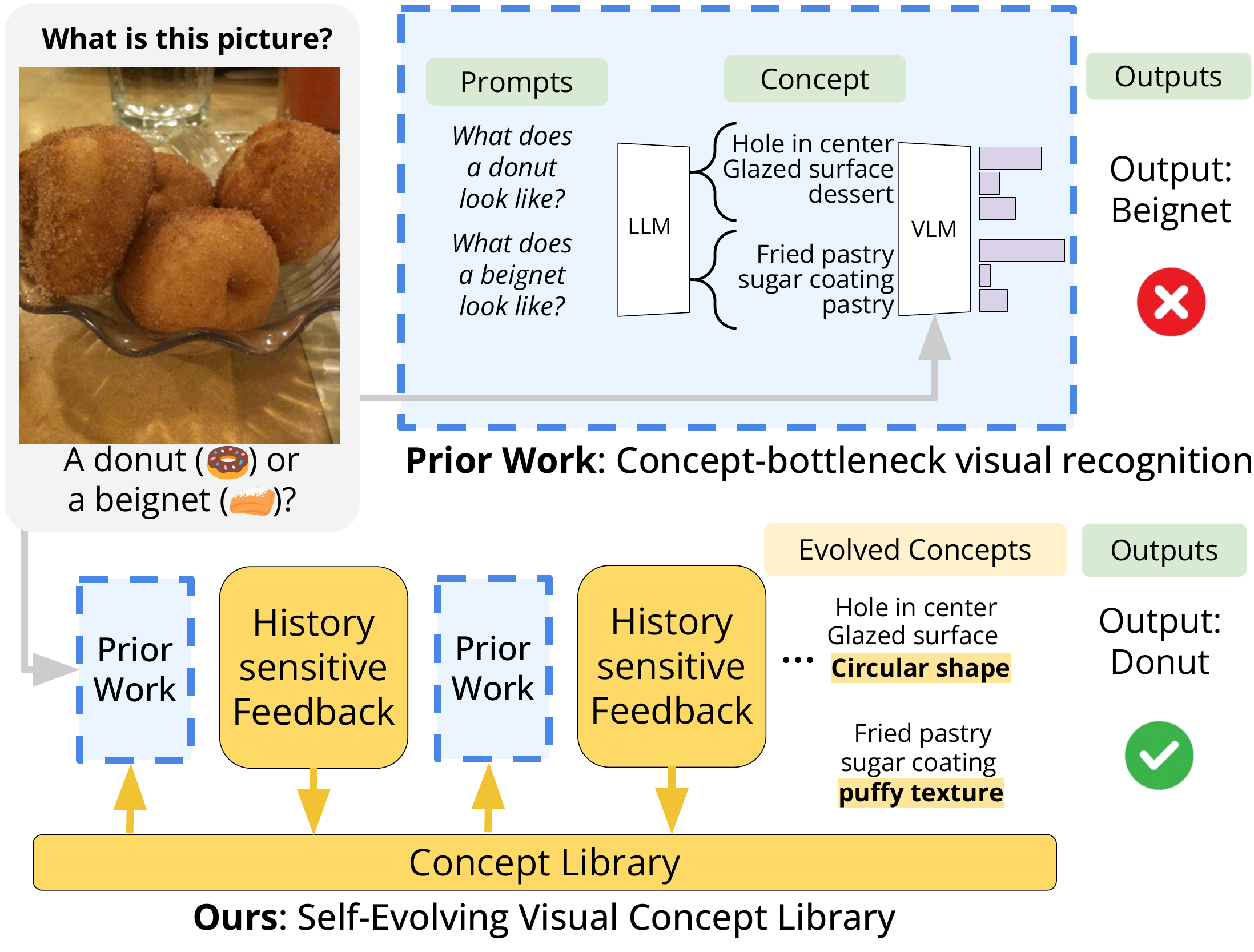}
    \caption{An overview of \name. Prior work: concept-bottleneck visual recognition aims to leverage discriminative visual concepts to enable more accurate object classification. Ours: \name is an approach for iteratively evolving a visual concept library using feedback from a VLM critic, to discover more effective visual concepts.
    }
    \label{fig:overview}
\end{figure}

How do humans recognize different visual categories? Consider the example in Figure~\ref{fig:overview}: while easily recognizable as a pastry, distinguishing between a ``donut'' and a ``beignet'' requires understanding visual concepts such circular shape with a hole for the donut, puffy texture, and the presence of powdered sugar specifically on the beignet. These visual concepts, including shape, texture, and the presence or absence of specific features, enable us to make distinctions between objects. Concept-bottleneck visual recognition~\cite{DBLP:conf/iclr/MenonV23,yang2023languagebottlelanguagemodel,yan2023learning} aims to leverage these discriminative visual concepts, to enable vision systems to more accurately recognize a wider range of classes. 
Here, we study general approaches for improving concept-bottleneck visual recognition systems by evolving visual concept libraries to find more effective concepts.

Existing concept-bottleneck visual recognition systems typically leverage a Large Language Model (LLM) to generate a set of potential visual concepts relevant to the task, then use a Vision-Language Model (VLM) to make predictions from these concepts. This process potentially improves both interpretability and accuracy for classification~\cite{DBLP:conf/iclr/MenonV23,yang2023languagebottlelanguagemodel,yan2023learning}. However, existing methods face limitations: manually defined concepts are labor-intensive, and LLM-generated concepts can be inaccurate or fail to account for interactions between them. We need more effective methods to construct and refine visual concepts. 
One promising approach to improve visual concept learning is to leverage library learning \cite{ellis2021dreamcoder,hu2024scenecraft,grayeli2024symbolic}, which focuses on building a reusable collection of components. While library learning has shown success in domains that are naturally symbolically decomposable (e.g., equation learning \cite{grayeli2024symbolic}), it is not well-explored for visual concept learning. Our key insight is that library learning complements visual concept learning by providing a structured and evolving repository of concepts that is more effective for visual recognition.

To achieve this, we introduce \escher, a novel self-evolving framework to automatically discover and refine a library of visual concepts. \escher employs an iterative algorithm where a VLM acts as a critic, providing feedback on the effectiveness of concepts generated by an LLM. Specifically, the VLM evaluates the similarity between each image and its associated concepts compared to other images. This evaluation, captured in a contrastive score, serves as a feedback signal to guide the LLM in refining its generated concepts. Furthermore, \escher provides the history of concepts and feedbacks to the LLM, enabling the LLM to effectively learn from its past performance and improving its proposals over time. Through this iterative process, \escher produces a set of concepts that are both accurate and highly informative for the VLM, enabling it to make more effective predictions.

Our approach offers several key advantages. First, it is broadly applicable and complements a range of existing concept-bottleneck visual recognition frameworks, including those designed for zero-shot, few-shot, and fine-tuned settings. This adaptability ensures that as LLMs, VLMs, and visual concept learning frameworks continue to evolve, \escher remains relevant and applicable to emerging techniques. Second, \escher requires no human annotations or labeled datasets, making it a plug-and-play solution for various visual recognition tasks. Finally, the iterative approach of \escher leverages the in-context learning capabilities of LLMs, allowing them to learn from their concept history and generate increasingly effective concept concepts. This iterative refinement process ensures that the concept library continuously adapts and improves, leading to more accurate and discriminative visual representations.

To summarize, our contributions are:
\begin{itemize}
    \item We present \escher, a novel VLM- and LLM-based framework for self-evolving visual concept libraries. Our method does not require human-labeled data and can improve the quality of the learned concepts via an open-ended learning loop.
    \item We develop an iterative concept refinement algorithm, leveraging the both the ability of VLMs to act as a critic and the ability of LLMs to incorporate history, to improve visual concepts based on past performance.
    \item We demonstrate that \escher is complementary to a range of different state-of-the-art baselines, and our learned concept library improves performance across zero-shot, few-shot, and fine-tuned image classification settings.
\end{itemize}

\begin{figure*}[ht]
    \centering
    \includegraphics[width=\linewidth]{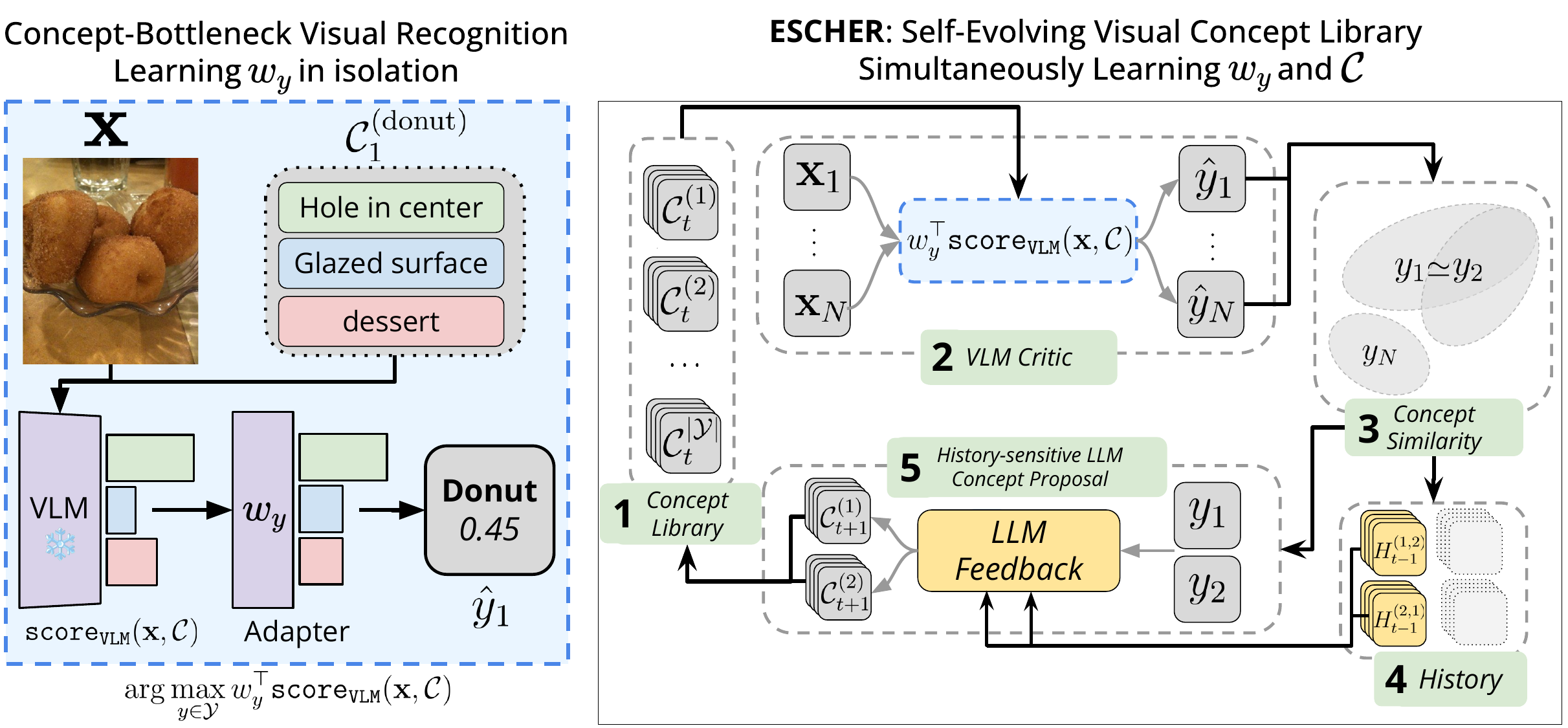}
    \caption{(\textbf{Left}) Existing work on concept-bottleneck visual recognition, where a VLM scores a set of concepts to perform classification. The classification is based on the class with the maximum concept scores. (\textbf{Right}) \name. (\textbf{1}) \name follows previous work \citep{DBLP:conf/iclr/MenonV23} in instantiating a set of concepts for each class using an LLM. (\textbf{2}) It initializes a concept-bottleneck model and collect the predictions for a classification dataset $\mathcal{D} = \{(\mathbf{x}_i, y_i)\}_{i=1}^N$ (labels optional). (\textbf{3}) A concept similarity heuristic identifies frequently confused classes. (\textbf{4}) A history bank then stores relevant information to guide (\textbf{5}) the LLM sampling procedure for improved concepts that disambiguate these classes. The new concepts are integrated into the next iteration.}
    \label{fig:architecture}
\end{figure*}

\section{Related Work}\label{sec:related_work}

\paragraph{Vision-language models.} VLMs have emerged as powerful tools for a wide range of visual tasks, from visual question answering to zero-shot image classification~\cite{radford2021learning,beyer2024paligemma,liu2024visual}. These models are typically trained on large datasets of image-text pairs, enabling them to reason about the relationship between these modalities. Some VLMs~\cite{jia2021scaling,radford2021learning,yu2022coca} are trained using contrastive learning objectives to learn embedding spaces aligned between image and text representations. These models have shown promising results for zero-shot and few-shot classification.

While VLMs have shown remarkable capabilities, they have limitations that motivate exploring alternative approaches, such as concept-bottleneck models. One limitation is their difficulty in perceiving and reasoning about fine-grained visual concepts. For example, recent work has shown that VLMs can struggle to distinguish between subtle visual differences \cite{tong2024eyeswideshutexploring}. 
As such, despite recent advances, these models can still often fail with accurate visual recognition. 
The standard zero-shot method does not provide any intermediate understanding or explanation of the model's reasoning process \cite{DBLP:conf/iclr/MenonV23}. These challenges highlight the need for more interpretable and controllable approaches to visual recognition, such as those offered by concept-bottleneck methods.

\paragraph{Concept-bottleneck Visual Recognition Models.}  Our work builds upon recent work on concept-bottleneck models \cite{saha2024improved, yang2023languagebottlelanguagemodel, yan2023learning, DBLP:conf/iclr/MenonV23, zhang2023prompt, Pratt2022WhatDA}, which first identify relevant concepts using a vision language model (VLM), and then uses those concepts to make a prediction.  Compared to directly querying a VLM, this approach has advantages including interpretability, (if the concepts are interpretable) and stronger accuracy (if the concepts are useful in capturing the classification task).  Similar ideas predate the rise of VLMs  \cite{li2018deep,kim2021xprotonet}, where a so-called concept bottleneck is built into neural network architectures, and concepts are learned via end-to-end training. Recent algorithms for learning concept-bottleneck models with VLMs generally fall within two categories:

\textit{Non-parametric Algorithms}: These methods focus on improving the visual concepts by employing non-parametric optimization techniques. One increasingly common approach is to use zero-shot queries to an LLM to select a list of concepts for each class that is useful for classification \cite{DBLP:conf/iclr/MenonV23}. The aggregated score for a class is the mean over the scores of the selected concepts. ~\cite{chiquier2024evolving} follows the LLM induction paradigm to initialize class concepts but, borrowing ideas from genetic mutation, repeatedly queries a finetuned LLM to generate new concepts and new concept selections for each class -- using binary classification loss to rate each mutation. This mutation process focuses on each class in isolation and must be repeated for each class, which proves to be impractical for datasets with more than 20 classes. \name, instead, reasons jointly about all classes and focuses on only those classes that are underperforming. As \name is agnostic to the choice of VLM and the number of samples needed for training, it is technically possible to integrate the llm-mutate framework within \name to maximize classification performance.

\textit{Parametric Algorithms}: These methods focus on improving the performance of a visual concept bottleneck classifier by training parametric adapters on top of the scores output by a frozen VLM model. These models may also subsample concepts from the concept library.  These architectures generally consist of a `concept bottleneck' \cite{yang2023languagebottlelanguagemodel,yan2023learning}, with additional learning required, such as a linear probing adapter or additional finetuning \citep{saha2024improved}. %
\name is agnostic to the choice of concept tuning method, and focuses on using such methods to guide the concept discovery loop. One can `plug' \name into any of these other works and observe a performance improvement while retaining the key characteristics of the type of architecture used.

\paragraph{Library Learning.} Library learning is an emerging direction of program synthesis that aims to automate the construction of reusable components (libraries) for program generation. This is often framed as a hierarchical Bayesian optimization problem, where the goal is to simultaneously learn the library of components and the optimal way to combine them to solve a given task \cite{ellis2021dreamcoder,lake2015human, grand2024lilo, Wong2021LeveragingLT}. While library learning has shown promise in domains like equation learning \cite{grayeli2024symbolic},
its application to visual concept learning presents unique challenges.

Unlike domains that are naturally symbolically decomposable, visual concepts often exhibit complex and subtle relationships that are difficult to capture with traditional library learning techniques. Moreover, the space of potential visual concepts is vast and diverse, making it challenging to design effective search strategies. Our work addresses these challenges by introducing a novel library learning approach designed for visual concept discovery. By leveraging a VLM as a critic and incorporating class resolution history into the LLM, our method can effectively explore the space of visual concepts and construct a dynamic library that adapts to the specific needs of the visual recognition task.

\section{Problem Formulation}\label{sec:problem_statement}

\paragraph{Concept-Based Visual Recognition}  Our work is rooted in the emerging area of concept-bottleneck visual recognition \cite{DBLP:conf/iclr/MenonV23,yang2023languagebottlelanguagemodel,yan2023learning} (Figure~\ref{fig:architecture}). Given an image $\mathbf{x}$, a set of concepts $\Con$, and a label $y$, the basic setup is to use a vision language model (VLM) to score the likelihood of each concept $c\in\Con$ for image $\mathbf{x}$, denoted by $\score(\mathbf{x},c)$. We denote the vector of scores over all concepts as $\score(\mathbf{x},\Con)$ . Subsequently, the aggregate score of a label $y$ for image $\mathbf{x}$ is a weighted sum over the concept scores: $f(\mathbf{x},y) = w_y^T\score(\mathbf{x},\Con)$, where $w_y$ can either be a learned or fixed parameter vector.\footnote{Some prior work uses a uniform vector for $w_y$ \cite{DBLP:conf/iclr/MenonV23}.}  Finally, classification over a fixed label set $\mathcal{Y}$ is performed by choosing the class that maximizes image-concept similarity.
\begin{align}\label{eq:cbm-inference}
    y^\star &= \arg\max_{y \in \mathcal{Y}} w^\top_{y} \texttt{score}_{\texttt{VLM}}(\mathbf{x}, \mathcal{C})
\end{align}

This setup encompasses the bulk of recent work in concept-bottleneck models for image classification \citep{DBLP:conf/iclr/MenonV23,yang2023languagebottlelanguagemodel,yan2023learning} and helps considerably in fine-grained and out-of-distribution classification scenarios. For instance, while a class $y=\text{``SpaceX Starship''}$ may not have been seen during CLIP training, we can construct a reasonably accurate and interpretable rocket classifier by aggregating the likelihoods over the feature set: $\{f(\mathbf{x}| c_y)~|~c_y \in [\text{`Stainless Steel Rocket'}, \text{`Grid Fins'}, \text{`Space X logo'}]\}$. Selecting this feature set can be automated using a foundation model with access to an external database \cite{schick2023toolformerlanguagemodelsteach}.

Typically, a concept-bottleneck model is developed by predefining a fixed set of concepts $\Con$, and then finetuning the weight matrix $\Adapter$ for each class. The optimization objective is then chosen to maximize the performance of the model and identify an interpretable set of concepts. A Bayesian formulation is presented in Eq~3.

In contrast, scientists -- when confronted with a new domain -- rarely rely on a static set of concepts.  The first reaction of a scientist is to learn more about the domain and expand their conceptual knowledge base. This newly gained knowledge is subsequently used to structurally discriminate between classes. For instance, even a trained ecologist might struggle to differentiate a Northern Curly-tailed Lizard from a Florida Scrub Lizard due to lack of prior knowledge, whereas a herpetologist can rely on their knowledge of lizard physiology to identify the correct characteristic feature difference \footnote{\url{https://www.inaturalist.org/observations/1970016}}

\paragraph{Visual Recognition with Latent Concept Libraries.} We model this evolving set of concepts as textual descriptions sampled from a latent concept library. We frame the relationship between the latent concept library and the classification model as a Hierarchical Bayesian model consisting of
(i) a prior $p(\Con)$ representing the natural distribution over concept libraries;
(ii) a model $p_{\mathcal{C}}(\Adapter| \Con)$ that quantifies the likelihood of assigning open-vocabulary classes for a given concept library; and
(iii) an evaluation function $p(\Data | \Con) := \argmax_{y \in \mathcal{Y}} w_y^T\score(\mathbf{x},\Con)$ which grounds the performance of a concept library using an image classification dataset ($\mathcal{D}$) and a VLM-based recognition engine. We assume that the distributions $p(\Con)$ and $p(\Adapter | \Con)$ can be approximated using LLMs. That is, we can prompt an LLM to generate interesting concepts, and we can prompt an LLM to generate and discover new concepts that adhere to an open-vocabulary category. We also assume that the VLM-based visual recognition engine is well calibrated for confidence estimation. We now pose the problem of visual recognition with latent concept libraries as one of simultaneously inducing an optimal set of concepts and an optimal concept-bottleneck visual reasoning model:
\begin{align}\label{eq:bayesian-formulation}
\Adapter^\star, \Con^\star &= \arg\max_{\Adapter, \Con} p(\Adapter, \Con | \Data) \\\nonumber
&= \arg\max_{\Adapter, \Con} \underbrace{p(\Data | \Adapter)}_{\text{CBM training}} \cdot \underbrace{p(\Adapter | \Con)}_{\text{By LLM}} \cdot \underbrace{p(\Con)}_{\text{By LLM}}
\end{align}

\section{Method}\label{sec:method}
\name performs a two stage evolution over the natural-language concepts and the weight matrix assignment. The two stages follow an alternating maximization strategy, as illustrated in Figure \ref{fig:architecture}: 
(1) \textit{Concept Bottleneck Optimization}: We fix a set of concepts and learn a concept-bottleneck model that maximizes the fitness to the dataset (Fig. \ref{fig:architecture}, {Left}). 
(2) \textit{History-sensitive concept evolution}: We leverage the best model to identify classes that appear to be confused and sample new concepts to resolve the confusion~(Fig. \ref{fig:architecture}, {Right}).

The rest of this section first describes the classical concept-bottleneck model maximization strategies. Then, we discuss common heuristics for identifying confused classes. Finally, we show how the classes can be disambiguated by sampling new concepts conditioned on feedback derived from previous evolutions.

\renewcommand*\Call[2]{\textproc{#1}(#2)}
\algrenewcommand\algorithmicindent{0.8em}
\newcommand{\AdapterSM}{w}

\begin{algorithm}
\small
\caption{Pseudocode for \name. \name takes as input a set of open vocabulary categories $\mathcal{Y}$, a dataset of images $\mathcal{D}$ (labels optional), a pretrained vision-language model $\Model$, an optional adapter  $\AdapterSM$ (We drop the $\cdot_\mathcal{Y}$ subscript for readability), and three hyperparameters: the number of iterations $T$, the decay rate $\gamma$ for repeated categories, and the number of pairs to evolve $K$. \name outputs two artifacts: the adapter parameters after $T$ iterations, $\AdapterSM_{T}$, and the corresponding evolved library of interpretable concepts $\Con_T$.}

\label{alg:main}

\begin{algorithmic}[1]
\Function{\name}{$\mathcal{Y}, \mathcal{D} = \{(\mathbf{x}_i, \mathbf{y}_i)\}_{i=1}^N, \theta_{\text{VLM}}, T, \gamma, K$}
    \State $\Con_0 \gets \textproc{InitConcepts}(\mathcal{Y})$ \Comment{Initialize concepts via zero-shot LLM queries}
    \State $H_0^{(i, j)} \gets \textproc{InitHistory}(\mathcal{Y}, T)$ \Comment{Track concepts and feedback for each class pair per iteration}
    
    \For{$t$ \textbf{in} range($T$)}
        \State $\AdapterSM_{t}^{\star} \gets  \textbf{fit}(\AdapterSM_t, \Con_t, \Data, \Model)$
        \State $\hat{\mathbf{y}} \gets \textbf{evaluate}(\Data, \Model, \AdapterSM_{t}^{\star})$
        \State $\{r_{ij}\}_{i,j=1}^{|\mathcal{Y}|^2} \gets \textproc{CalculateSimilarity}(\hat{\mathbf{y}})$
        
        \State $H_{t}^{(i,j)} \gets \textproc{UpdateHistory}(\{r_{ij}\})$ \Comment{Store the similarity of $y_i$ and $y_j$ for iteration $t$}
        \State $s_{ij} \gets \textproc{ComputeSampleProb}(r_{ij}, H^{(i,j)}_{[1:t]}, \gamma)$
        \State $\{r_{ij}\}_{1}^{K} \gets \textbf{subsample}(\{r_{ij}\}, s_{ij})$
        
        \For{\textbf{each} $(i, j) \in \{r_{ij}\}^K_1$}
            \State $\hat{c}^{(i)}, \hat{c}^{(j)} \gets \textproc{ConceptEvol}(y_i, y_j, \Con_t^{(i)}, \Con_t^{(j)}, H^{(i,j)}_{[1:t]})$
            \State $\Con_{t+1}^{(i)} \gets \Con_{t}^{(i)} \cup \{\hat{c}^{(i)}\}$
            \State $\Con_{t+1}^{(j)} \gets \Con_{t}^{(j)} \cup \{\hat{c}^{(j)}\}$
            \State $H^{(i,j)}_{t+1} \gets \textproc{UpdateHistory}(\Con_{t+1}^{(i)}, \Con_{t+1}^{(j)})$ \Comment{Store the updated concepts for iteration $t+1$}
        \EndFor
        
        \State $\AdapterSM_{t+1} \gets \AdapterSM_{t}^{\star}$
        \State $C_{t+1} \gets C_t$
    \EndFor
    
    \State \Return $\AdapterSM_T, \Con_T$
\EndFunction
\end{algorithmic}
\end{algorithm}

\paragraph{Concept Bottleneck Optimization.} Concept-bottleneck models \cite{koh2020conceptbottleneckmodels} generate their predictions by learning to linearly combine the intermediate predictions over a fixed set of interpretable concepts. This yields a high-performing yet interpretable classifier that can be used for downstream classification tasks. We focus on two paradigms within this field for optimizing the adapter weight matrix.

\textit{Zero shot maximization.} In this setting, the adapter weights are instantiated by an LLM. Intuitively, the adapter will take the form of a block diagonal matrix, where each block represents the concepts selected by the LLM for a particular class and each element in the block is assigned the uniform weight $1/{|c_y|}$, where $c_y$ is the set of concepts the LLM generates for a particular label $y$. As no labels are needed, this paradigm generates extremely flexible classifiers. However, the efficacy of the concepts is deeply tied to the backbone VLM's ability to score fine-grained concepts. 

\textit{Few shot / Fine-tuned maximization.} Under this paradigm, the adapter weights are instantiated as a learnable linear layer of shape $\mathbb{{R}^{|\Con| \times |\mathcal{Y}| }}$. As the set of concepts can grow very large, most approaches subsample the set of concepts as well. The linear layer is trained with cross-entropy loss, often with various regularizers \citep{yan2023learning}. In the few-shot setting, the number of images per class is fixed while in the fine-tuned setting, no restriction is placed on the number of images. This adapter training approach also overcomes the inherent weakness of the zero shot setting, as concepts that do not add to the performance of the model can be down-weighted or ignored by the linear adapter. 

\paragraph{Heuristics for disambiguation.}
After maximizing the model for the given set of concepts, the \textsc{CalculateSimilarity} function identifies classes that are confused with each other. This function takes as input a matrix of scores for the dataset of images $\hat{\mathbf{y}} \in \mathbb{R}^{N\times |\mathcal{Y}| }$. Each value in $\hat{\mathbf{y}}$ indicates the image-class similarity. We leverage this matrix of scores to identify classes that are confused with each other across the images in the dataset. $\{r_{ij}\}_{i,j=1}^{|\mathcal{Y}|^2}$ denotes the list of all possible confusion scores. \name's modular framework allows us to admit a wide variety of heuristics to model this confusion score. We list several heuristics to measure this disambiguation signal that perform well empirically. Additional studies are presented in \S~8.4, Figure~4, and Figure ~5.

\begin{itemize}
    \item \textit{Top-k Confusion} Top-k confusion is computed by sorting each row in descending order of score and keeping the top $k$ class predictions. That is: $r_{ij} = \text{confusion\_mat}(\text{sort}(\hat{\mathbf{y}})[:, :k])$. Consult Figure~4 for more insight into how this heuristic works.
    \item \textit{Correlation}: We compute Pearson's correlation between each class $r_{ij} = \text{corrcoef}(\hat{\mathbf{y}}^\top)$. Pearson's correlation measures linear correlation between two classes ($y_i$ and $y_j$) with -1 denoting maximal  negative correlation and 1 denoting maximal positive correlation. This is visualized in Figure~5. The hyperparameter $k$ differs from the hyperparameter $K$ which tracks the number of classes to disambiguate.
    \item \textit{Agglomerative Clustering}: Identifying classes with similar response signals can be viewed as an unsupervised clustering problem over the columns of $\hat{\mathbf{y}}$. The number of clusters is kept as a tunable hyperparameter, and we choose the values of ${r_{ij}}$ by greedily bottom-up traversing the dendrogram.  
    \item \textit{Confusion Matrix}: \name assumes no access to labels during evolution. However, if labels are provided, we can construct a confusion matrix to identify classes that are confused for each other.
\end{itemize}

\paragraph{History-sensitive concept evolution.} 
Each value in $\{r_{ij}\}_1^K$ represents two classes that the model is unable to discriminate between. This suggests that the model lacks the correct discriminative concepts in the concept space $\Con$ to conceptually separate $y_i$ and $y_j$.  Now we use the \textsc{ConceptEvolution} function, which uses a zero-shot prompt to extract a list of natural language concepts for both classes ($\hat{c}_i, \hat{c}_j$) that add additional concepts useful for disambiguating $y_i$ and $y_j$ into the concept library. 

This task requires strong reasoning skills, so we provide the model with a scratchpad to enhance the model's reasoning abilities \cite{nye2021workscratchpadsintermediatecomputation}. More details are presented in \S~8.5.

This process is repeated for each pair of confused concepts. However, the $\textproc{CalculateSimilarity}$ function generates $|\mathcal{Y}|^2$ pairs in each iteration, and processing all such pairs is extremely inefficient for real-world datasets. To increase feedback generation efficiency, we only sample the top $K$ pairs from a random distribution weighted by the confusion coefficient score and a penalty for repeat confusions which increases exponentially, controlled by a decay parameter $\gamma$. This is implemented as $\textproc{ComputeSampleProb}(r_{ij}, H_{[1:t]}^{(i,j)}, \gamma)$ where the size of $H_{[1:t]}^{(i, j)}$ indicates the number of times the $(i, j)^{\text{th}}$ classes have been confused for each other. The $\gamma$ parameter helps guard against classes that are often misclassified due to failures in the backbone ViT model.

Additionally, each pairwise disambiguation can cause collisions with other classes in later iterations, which makes it likely that some class pair will need multiple rounds of feedback. If we generate feedback in isolation, the model is likely to regurgitate previously proposed features. To ensure that each new round of feedback generates novel and interesting concepts, we borrow ideas from the program synthesis literature \citep{austin2021programsynthesislargelanguage} and append past `execution traces' to the model query. This history-conditioned prompt is presented in Figure~6. We maintain a history of past evolutions for each pair $(y_i, y_j)$ along with the similarity score derived from the VLM critic's score of the modified concepts $c_i \cup \hat{c}_i$ and $c_j \cup \hat{c}_j$. Concretely, \textsc{InitializeHistory} instantiates this datastructre, and \textsc{UpdateHistory} updates the relevant fields in each iteration (the new concepts added to class $i$ and $j$ in iteration $t$ and the class confusion score in iteration $t+1$ after incorporating feedback). These functions are explored in more detail in \S~8.10. 

\section{Experiments}\label{sec:experiments}

\name is a meta-algorithm that aims to enhance the performance of concept-bottlenecked models (CBMs) by learning a library of concepts using an alternating maximization loop. Such CBMs operate in diverse data regimes, with some models requiring no human labels \cite{DBLP:conf/iclr/MenonV23} and others requiring a fully annotated classification dataset  (\cite{yan2023learning}). Our experiments focus on studying whether \name can improve the performance of such pre-existing algorithms that are representative of their respective data regime they operate in, with no additional modifications (\S \ref{sec:expts.finetuned}, \S \ref{sec:expts.fewshot}, and \S \ref{sec:expts.zeroshot})

In addition, \S\ref{sec:expts.library-ablation} presents an ablation of \name's library learning component, and \S\ref{sec:expts.backbone-ablations} explores \name's performance under various LLM and VLM backbones. More studies are presented in the Appendix (\S~8).

\paragraph{Datasets.} We demonstrate the effectiveness of \name on seven fine-grained and general purpose classification datasets that generally fall into three categories: finegrained classification datasets with scientific applications (NABirds~\cite{van2015building} and CUB~\cite{wah2011caltech}), fine-grained classification datasets with general purpose applications (Food-101~\cite{bossard2014food}, Stanford Cars~\cite{krause20133d}, Flowers-102~\cite{nilsback2008automated}), and general purpose categorization datasets (CIFAR100~\cite{krizhevsky2009learning}). 

\paragraph{Evaluation.} We extend three concept-bottleneck visual classification models with reproducible, publicly available codebases at the time of writing: LM4CV (finetuned adapter), LaBO (few shot adapter), and  Classify by descriptions (zero-shot adapter) \citep{yan2023learning, yang2023languagebottlelanguagemodel, DBLP:conf/iclr/MenonV23}. We follow previous work in using Top-1 accuracy to evaluate the performance of the CBM model on the test set of the classification dataset.

\paragraph{Methodology.} All experiments begin with an initial set of concepts generated with a backbone LLM (\texttt{gpt-3.5-turbo-0125} \citep{brown2020language}). A CBM conditioned on these concepts produces logits for the images in the validation set, and a heuristic identifies classes that are confused for each other. The specific heuristic and hyperparameters vary depending on the task and underlying algorithm, and grid-searching for such hyperparameters proves to be practical and effective. We use a generic history-conditioned prompt presented in Fig.~6. We use a ViT-L/14 CLIP model \citep{radford2021learning} as the vision backbone for all experiments unless otherwise clarified.

\subsection{Comparison against fine-tuned baselines}\label{sec:expts.finetuned}
\begin{table}[htbp]
\centering
\begin{tabular}{lcc}
\toprule
& LM4CV & LM4CV+\name \\
\midrule
CIFAR-100 & 84.48 & \textbf{89.63} \\ 
CUB-200-2011 & 63.26 & \textbf{83.17} \\ 
Food101 & 94.77 & \textbf{94.90} \\ 
NABirds & 76.58 & \textbf{78.21} \\ 
Oxford Flowers & 94.80 & \textbf{96.86} \\ 
Oxford IIIT Pets & 92.50 & \textbf{92.86} \\ 
Stanford Cars & 86.84 & \textbf{93.76} \\ 
\bottomrule
\end{tabular}
\caption{Top-1 accuracy of LM4CV \cite{yan2023learning} and LM4CV evolved with \name on multiple fine-grained classification problems. \name improves upon LM4CV's performance in all datasets while utilizing no extra human annotations.}
\label{tab:finetune-performance}
\end{table}

\paragraph{Setup.} For each dataset, we generate an initial set of concepts using queries to \texttt{gpt-3.5-turbo} and a generic visual concept learning prompt adapted from \citep{DBLP:conf/iclr/MenonV23}. We use LM4CV's suggested hyperparameters wherever possible. We run 60 iterations of \name. Similarity is computed using the Top-k confusion metric ($k=3$), only the top 50 pairwise confusions are evolved, and decay rate $\gamma$ is set to $1/30$ (i.e.: after 30 repeated evolution calls, the value drops to half of the original). LM4CV and LM4CV+Escher are trained for the same number of steps per iteration with the same batch size and learning rate. 

\paragraph{Observations.} Our observations are presented in Table \ref{tab:finetune-performance}. We draw three observations from this experiment. First, LM4CV + \name achieves a higher Top-1 accuracy than vanilla LM4CV on all datasets, suggesting that learning a concept library is an effective axis of improvement. Second, LM4CV+\name significantly improves the performance of LM4CV on datasets where the initial accuracy is low, and finally, LM4CV+\name did not require any human provided information to achieve this result. 

\subsection{Comparison against few-shot baselines}\label{sec:expts.fewshot}

\begin{table}[htbp]
\centering
\adjustbox{max width=0.45\textwidth}{
\begin{tabular}{l*{2}{cc}}
\toprule
& \multicolumn{2}{c}{8 shot} & \multicolumn{2}{c}{16 shot} \\
\cmidrule(lr){2-3} \cmidrule(lr){4-5}
Dataset & LaBO & LaBO+\name & LaBO & LaBO+\name \\
\midrule
CIFAR100 & \textbf{74.23} & 73.62 & \textbf{77.67} & 77.23 \\ 
CUB & 72.78 & \textbf{73.37} & 78.75 & \textbf{78.79} \\ 
Food101 & \textbf{87.02} & 86.10 & 88.49 & \textbf{88.50} \\ 
Oxford Flowers & \textbf{95.66} & 95.37 & 97.69 & \textbf{97.80} \\ 
Stanford Cars & 75.07 & \textbf{75.99} & 81.48 & \textbf{82.56} \\ 
\bottomrule
\end{tabular}}
\caption{Top-1 accuracy for LaBO and LaBO evolved with \name on multiple fine-grained classification datasets in a few-shot learning setting. LaBO benefits from \name's library guidance in the 16 shot setting more than in the 8 shot setting.  We observe mixed results on \name's efficacy in the few shot domain.}
\label{tab:few-shot-performance}
\end{table}

\paragraph{Setup.} Our setup for LaBO follows the same setup as that of LM4CV. We evaluate all datasets for 8 shot and 16 shot. This setup necessitates a balanced set of images for each class. We drop NABirds from this evaluation, as some NABirds classes contain as low as 3 images per class. We keep the same hyperparameters as our LM4CV experiments but do not use a decay rate as the number of repeat classifications is generally unproblematic. 

\paragraph{Observations.} We observe mixed results in few-shot evaluation, with similar performance compared to the baseline on all datasets except for Stanford Cars, which showed modest improvement, and CIFAR100, which showed very modest deterioration (Table \ref{tab:few-shot-performance}). In general, LaBo+\name performs considerably better in the 16-shot setting compared to the 8-shot setting. We hypothesize these mixed results are induced by poor calibration of LaBO's CBM's and as a result of model overfitting on the few available labels for each class.

\subsection{Comparison against zero-shot baselines}\label{sec:expts.zeroshot}

\begin{table}[htbp]
\centering
\begin{tabular}{l|c|cccc}
\toprule
Dataset & CLIP & CbD & CbD+\name \\
\midrule
CIFAR-100 & 73.30 & 76.20 & \textbf{77.80} \\
CUB-200-2011 & 64.83 & 62.00 & \textbf{63.33} \\
Food101 & 92.51 & 93.11 & \textbf{93.58} \\
NABirds & 53.53 & 53.61 & \textbf{54.30} \\
Oxford Flowers & 74.51 & 79.41 & \textbf{81.37} \\
Stanford Cars & 74.53 & 75.65 & \textbf{77.14} \\
\bottomrule
\end{tabular}
\caption{Top-1 accuracy of CLIP (ViT-L/14) \cite{radford2021learning}, Classify by Descriptions (CbD) \cite{DBLP:conf/iclr/MenonV23}, and CbD evolved with \name on multiple fine-grained classification datasets in a zero-shot learning setting. CbD+\name improves upon CbD's performance in all datasets.}

\label{tab:zeroshot-performance}
\end{table}

\paragraph{Setup.} Our zero shot setup compares against Classify by Descriptions. We sample all concepts from \texttt{gpt-3.5-turbo}. We use the same hyperparameters as LM4CV experiments, except for setting the decay rate to 50. We continue to use the ViT-L-14 backbone mode.

\paragraph{Observations.} Our observations are presented in Table \ref{tab:zeroshot-performance}. Overall, we find that CbD's performance improves consistently for all datasets when we evolve the concepts with \name. The relative improvement in performance is less than that in LM4CV as the backbone  (CLIP ViT-L-14) model's output scores are less calibrated than those produced by the finetuned adapter, which leads to noisier iterations. 

\begin{figure*}[htbp]
    \centering
    \includegraphics[width=\linewidth]{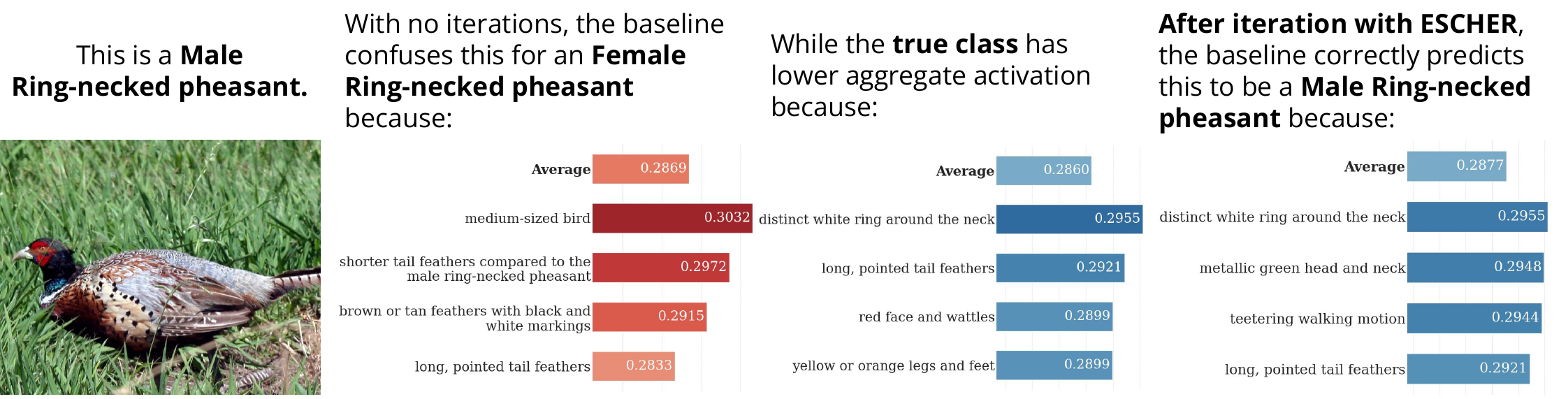}
    \caption{A qualitative example of evolving concepts with CbD+\name in NABirds. Initially, the model is confused between two similar categories with almost the same mean CLIP activation indicating that the concepts provide a coarse categorization signal, but miss subtle nuances. After training with \name, the feedback mechanism identifies new characteristic features (e.g. \textit{metallic green head and neck}) enabling the correct classification. Additional examples are provided in \S~8.6.}
    \label{fig:qualitative-results}
\end{figure*}

\subsection{Library Learning Ablation}\label{sec:expts.library-ablation}
\begin{table}[htbp]
\centering
\adjustbox{max width=0.45\textwidth}{
\begin{tabular}{lccc}
\toprule
Dataset & LM4CV & LM4CV & LM4CV \\
 &  & + Many Concepts & + \name \\
\midrule
CIFAR-100 & 84.48 & 86.91 & \textbf{89.63} \\
CUB-200-2011 & 63.26 & 66.09 & \textbf{83.17} \\
Food101 & 94.77 & 94.77 & \textbf{94.90} \\
NABirds & 76.58 & 76.28 & \textbf{78.21} \\
Oxford Flowers & 94.80 & 94.51 & \textbf{96.86} \\
Oxford IIIT Pets & 92.50 & 92.02 & \textbf{92.86} \\
Stanford Cars & 86.84 & 86.84 & \textbf{93.76} \\
\bottomrule
\end{tabular}
}
\caption{Top-1 accuracy of an ablation of \name's library learning component. For LM4CV, we replace the concepts learned with library learning with an equal number of concepts sampled from an LLM. We find that concepts evolved with \name still outperform naively sampling more concepts -- suggesting that feedback from a VLM critic is essential for LM4CV+\name's performance.}
\label{tab:ablation}
\end{table}

\name focuses on using zero shot disambiguation queries to maximize the performance of a concept library. To do this, \name makes asymmetrically more calls to a zero-shot LLM model than the baseline (which only samples concepts once). To verify that \name's performance is not simply a result of more concepts or greater number of LLM calls, we rerun LM4CV with three times more concepts per class than the initial set of concepts (for a total of 3810 concepts). These concepts are sampled with the same LLM backbone used in \name. This represents an ablation on the Library learning component used in \name.

Results are highlighted in Table~\ref{tab:ablation}. We find that LM4CV's performance does not significantly improve even given the same number of LLM sampled concepts as LM4CV+\escher. This suggests that (1) the library learning component is essential for concept evolution and (2) sampling concepts without incorporating feedback from the VLM underperforms integrating a VLM critic into the library learning loop.

We visualize qualitative results of concepts for each dataset and corresponding activated images in the supplementary material.

\subsection{Backbone VLM/LLM Ablation}\label{sec:expts.backbone-ablations}

\begin{table}[htbp]
\centering
\adjustbox{max width=0.45\textwidth}{
\begin{tabular}{lc|cc|cc}
\toprule
ViT-B/16 & CLIP & LM4CV & LM4CV & CbD & CbD \\
& & & +\name &  & +\name \\
\midrule
CIFAR-100       & 63.20 & 81.35 & \textbf{81.72} & 67.70 & \textbf{69.90} \\ 
CUB-200-2011    & 57.06 & 70.90 & \textbf{77.17} & 56.16 & \textbf{56.16} \\ 
Food101         & 87.24 & 92.00 & \textbf{92.20} & 88.51 & \textbf{89.19} \\ 
NABirds         & 44.56 & 66.69 & \textbf{68.71} & \textbf{45.25} & \textbf{45.25} \\ 
Stanford Cars   & 61.86 & 80.87 & \textbf{81.82} & 65.96 & \textbf{66.34} \\
\bottomrule
\end{tabular}
}
\caption{
Top-1 accuracy for evolving \name with a weaker LLM (Llama-3.3-70B-4bit) and visual critic (ViT-B/16). \name consistently improves the performance of LM4CV and CbD across datasets.
}
\label{tab:r.clipb16}
\end{table}

\name is a meta-algorithm that enhances the performance of existing CBMs. The performance of such models is inherently bottlenecked by the quality of their CLIP-based visual backbone for capturing concept-image relationships and the quality of the GPT-based language backbone for querying relevant concepts. In this experiment, we investigate whether \name can improve CBMs even with alternative VLM/LLM backbones. Concretely, we instantiate LM4CV and CbD with a new backbone LLM (4bit quantized \texttt{Llama-3.3-70B-Instruct} \citep{grattafiori2024llama}) and with new backbone VLMs (ViT-B/16 and ViT-B/32). These backbones are slightly weaker than the base models used in other experiments. As a result, we expect an overall reduction in performance and a weaker learning signal from the visual critic. We maintain the same training and evaluation setup as used in other experiments.

\paragraph{Observations.} Results for these experiments are presented in Table \ref{tab:r.clipb16} (for ViT-B/16) and in the Appendix Table~10 (for ViT-B/32). Overall, we observe that CBMs tend to perform better when paired with stronger backbones. Additionally, in every case, refining CBMs iteratively using \name leads to better performance than relying on a fixed set of concepts.

\section{Conclusion}\label{sec:conclusion}

We present \escher, a framework for evolving visual concept libraries for visual recognition. \escher iteratively updates the library using a VLM as a critic to guide an LLM to generate more effective concepts. This process also enables the LLM to incorporate past histories of the feedback from the VLM. Notably, \escher does not require any additional annotations, and is compatible with a range of concept-bottleneck visual recognition systems. We demonstrate our results on concept-bottleneck models in zero-shot, few-shot, as well as fine-tuned settings.

One direction of future work is to further improve the performance of evolving visual libraries on few-shot settings, by incorporating few-shot learning into the library evolution process and allowing the algorithm to leverage limited labeled data. 
Additionally, we aim to study the application of our approach to more complex visual reasoning tasks, where the learned concept libraries could provide a foundation for higher-level reasoning. 
By demonstrating the potential of combining library learning and concept-bottleneck visual recognition, we aim to encourage further research at this intersection, towards the development of more robust, interpretable, and intelligent visual recognition systems.

\clearpage
\paragraph{Acknowledgments}\label{sec:acknowledgements}
This research was partially supported by the NSF Expeditions award \#CCF-1918651, a DARPA award \#HR00112320018, an ARO award \#W911NF2110009, and a DARPA TIAMAT award \#HR0011-24-9-0431.

\clearpage
\setcounter{page}{1}
\maketitlesupplementary
\section{Appendix}\label{sec:appendix}

\name presents a modular framework for learning a library of visual concepts. In this section, we present additional discussions on the reasoning and validity behind several of \name's internal modules.
\begin{itemize}
    \item \S \ref{sec:a.history_ablation}: How does \name's performance change without history conditioning?
    \item \S \ref{sec:a.visual_critic_ablations}: How much impact does the visual critic have in \name's iterations?
    \item \S \ref{sec:a.multiple_critics}: How does \name perform with different disambiguation heuristics? 
    \item  \S \ref{sec:a.zero-shot-prompts}: \name's history-conditioned prompt. 
    \item  \S \ref{sec:a.qualitative_results}: Qualitative results of \name's learned library.
    \item  \S \ref{sec:a.eschers-speed}: Practical considerations for using \name.
    \item  \S \ref{sec:a.intuitive-liblearning-explanation}: \name's convergence properties.
    \item \S \ref{sec:bayesian-formulation-cbm} Bayesian formulation for CBMs.
    \item \S \ref{sec:a.additional-ablation} Additional backbone ablation with ViT-B/32.
\end{itemize}

\subsection{\name without history conditioning}\label{sec:a.history_ablation}

\name utilizes a history conditioning component to store pairs of concepts which have been disambiguated more than once. This pairwise history is provided in the input context to the model alongside the feedback from the visual critic. To understand the effectiveness of the history conditioning component, we conduct additional experiments on the LM4CV domain by ablating the history conditioning model. 

\paragraph{Setup.} We replace the history conditioned concept space with a simpler prompt that doesn't use the history conditioning. The overall structure of the original prompt is left unchanged, except for the removal of previous conversations as well as the removal of a line with feedback parsing instructions.

\paragraph{Results.} Results are showcased in Table \ref{tab:history_ablation}. We find that the history conditioning is useful in almost all cases other than the Food101 dataset and the Oxford Flowers102 dataset, where the history conditioning ablation achieves comparable accuracy to other model.

\begin{table}[htbp]
\centering
\begin{tabular}{lrrr}
\toprule
ViT-L-14 & LM4CV & Ours & Ours-H \\
\midrule
CIFAR-100 & 84.48 & \textbf{89.63} & 86.73 \\ 
CUB-200-2011 & 63.26 & \textbf{83.17} & 81.58 \\ 
Food101 & 94.77 & \textbf{94.97} & \textbf{94.97} \\ 
NABirds & 76.58 & \textbf{78.21} & 77.35 \\ 
Oxford Flowers & 94.80 & \textbf{96.86} & \textbf{96.86} \\ 
Stanford Cars & 86.84 & \textbf{93.76} & 88.09 \\ 

\bottomrule
\end{tabular}
\caption{Top-1 accuracy for LM4CV+\name with and without history conditioning. We find that history conditioning helps improve the model performance in the majority of the datasets.}
\label{tab:history_ablation}
\end{table}

\subsection{Visual Critic Ablations}\label{sec:a.visual_critic_ablations}

\name relies on a visual critic for classes that are confused for each other. In this section, we investigate whether the quality of the visual critic impacts the concepts generated by \name. 

\paragraph{Setup.} We consider two experiments to study the impact of the visual critic. First, we replace the score matrix $S$ with a randomly generated score matrix and use this uninformative matrix to sample new concepts. If \name relies on the VLM critic, we expect to observe considerable performance deterioration. Second, we collect logits from a well-calibrated VLM critic and qualitatively compare the true confusion matrix and the correlation matrix calculated with various heuristics. These matrices have the same shape ($\mathbb{R}^{|\mathcal{Y}| \times |\mathcal{Y}|}$) and we expect to see similar trends emerge (despite differing absolute values).

\paragraph{Results.} We showcase results in Table \ref{tab:visual_critic_ablations.1} and Figure \ref{fig:visual_critic_ablations.2}. We observe that without the visual critic, each iteration degenerates into a random search over disambiguation pairs. The search is not completely unguided, as the decay rate helps the model avoid repetitive queries. Yet, the ablation underperforms LM4CV+\name in all datasets. 

Furthermore, our qualitative study reveals that \name's heuristic closely aligns with the ground-truth confusion matrix, without using any labels. However, \name's heuristic is sensitive to small errors, leading to slightly suboptimal disambiguation.

\begin{table}[htbp]
\centering
\begin{tabular}{lrrr}
\toprule
ViT-L-14 & LM4CV & Ours & Ours-H-VC \\
\midrule
CIFAR-100 & 84.48 & \textbf{89.63} & 86.76 \\ 
CUB-200-2011 & 63.26 & \textbf{83.17} & 82.41 \\ 
Food101 & 94.77 & \textbf{94.97} & 94.95 \\ 
NABirds & 76.58 & \textbf{78.21} & 77.29 \\ 
Oxford Flowers & 94.80 & \textbf{96.86} & 96.08 \\ 
Stanford Cars & 86.84 & \textbf{93.76} & 88.12 \\ 
\bottomrule
\end{tabular}
\caption{Top-1 accuracy for LM4CV, LM4CV+\name, and LM4CV without a visual critic or history conditioning. We find that the visual critic improves model performance in all cases.}
\label{tab:visual_critic_ablations.1}
\end{table}

\subsection{\name with low quality initial libraries}\label{sec:a.initial_concepts}

\name primary objective is to augment a CBM training loop by using VLM feedback to grow a library of natural language concepts. We have demonstrated that \name enhances the performance of various algorithms under a fixed library of components. In this section, we explore \name's behavior when the initial library of concepts is small or incomplete. 

\paragraph{Setup.} We start with the original concepts  generated with \texttt{gpt-3.5-turbo} and randomly subsample to $25\%$ and $50\%$ of the original number of per-class concepts. We use CbD as the testing domain for this experiment.

\paragraph{Results.} Results are reported in Table \ref{tab:a.initial_concepts}. We find that \name is able to sufficiently recover its base performance and, surprisingly, in some cases the reduction in initial concepts allows \name to discover higher-quality concepts, increasing overall performance.

\begin{table}[htbp]
\centering

\adjustbox{max width = \linewidth}{
\begin{tabular}{lrrrrrr}
\toprule
ViT-L-14 & CbD & CbD 50\% & CbD 25\% & Ours & Ours 50\% & Ours 25\% \\
\midrule
CIFAR-100 & 76.20 & 73.50 & 67.50 & \textbf{77.80} & 77.80 & 78.00 \\ 
CUB-200-2011 & 62.00 & 60.83 & 62.50 & \textbf{63.33} & 63.00 & 63.17 \\ 
NABirds & 53.61 & 54.38 & 54.26 & 54.30 & 55.15 & \textbf{55.44} \\ 
Oxford Flowers & 79.41 & 76.47 & 72.55 & 81.37 & 80.39 & \textbf{83.33} \\ 
Stanford Cars & 75.65 & 75.28 & 74.91 & \textbf{77.14} & 76.77 & 76.52 \\ 
\bottomrule
\end{tabular}
}
\caption{Top-1 accuracy for CbD+\name with different qualities of initial libraries. We find that \name achieves the best performance for this dataset and, in some cases, benefits from less initial concepts.}
\label{tab:a.initial_concepts}
\end{table}

\subsection{Visual Concept learning with diverse critics} \label{sec:a.multiple_critics}

\name allows multiple ways of specifying the heuristic for  disambiguating label pairs. Each heuristic has its own advantages and shortcomings. In this experiment, we conduct a quantitative study of various heuristics and their impact on performance. 

\paragraph{Setup.} We target the finetuning experiments with LM4CV and train our model with three different disambiguation heuristics: \textit{PCA+Correlation} (PCA), \textit{Earth Mover's Distance} (EMD), and \textit{Pearson's Correlation} (PCC). We retrain LM4CV + \name using the same hyper parameters as used in the best performing \textit{Top-k Confusion} experiments.

\paragraph{Results.} Results are reported in Table \ref{tab:a.multiple_critics}. We find that \name performs best with \textit{Top-k Confusion}. This heuristic also offers fine-grained control over the sensitivity of which pairs should be disambiguated. Figure \ref{fig:visual_critic_ablations.2} and Figure \ref{fig:visual_critic_ablations.2.1} offer a qualitative analysis of the confusion matrices for three datasets.

\begin{table}
\centering
\adjustbox{max width=\linewidth}{
\begin{tabular}{lrrrrr}
\toprule
ViT-L-14 & Baseline & Ours & Ours & Ours & Ours \\
 &  &  & +PCA & +EMD & +PCC \\
\midrule
CIFAR-100 & 84.48 & \textbf{89.63} & 86.58 & 86.71 & 86.44 \\ 
CUB-200-2011 & 63.26 & \textbf{83.17} & 80.34 & 78.91 & 75.13 \\ 
Food101 & 94.77 & \textbf{94.97} & 94.82 & 94.89 & 94.77 \\ 
NABirds & 76.58 & \textbf{78.21} & 76.99 & 76.87 & 76.32 \\ 
Stanford Cars & 86.84 & \textbf{93.76} & 87.54 & 87.35 & 86.94 \\ 
\bottomrule
\end{tabular}
}
\caption{Top-1 accuracy for LM4CV+\name using various disambiguation heuristics. We find that \textit{Top-k Confusion} offers the most fine-grained control over the sensitivity of the disambiguation and performs the best empirically. }
\label{tab:a.multiple_critics}
\end{table}

\subsection{Zero shot prompts.}\label{sec:a.zero-shot-prompts}

We show the prompts we used to self-evolve the library in \name in Figure \ref{fig:zero-shot-prompts}. The prompts are designed to make use of a scratchpad. Specifically, when \name asks an LLM to disambiguate two classes (Fig.\ref{fig:zero-shot-prompts}), it first requests a reasoning for the proposed concept, then the concept itself in a separate JSON field. Without the additional reasoning field, the LLM often merges the explanation with the concept, producing a lengthy concept that exceeds the CLIP text encoder's context length and prematurely halts \name's iterative loop.

\subsection{Additional qualitative results}\label{sec:a.qualitative_results}

We highlight additional qualitative results of images and classes that become better separated because of \name's iteration process in Figure \ref{fig:qualitative-results.1}.

\begin{figure*}
    \centering
    \textbf{Qualitative analysis of `Pearson's Correlation' as a disambiguation heuristic.}\par\medskip
    \includegraphics[height=0.25\paperheight]{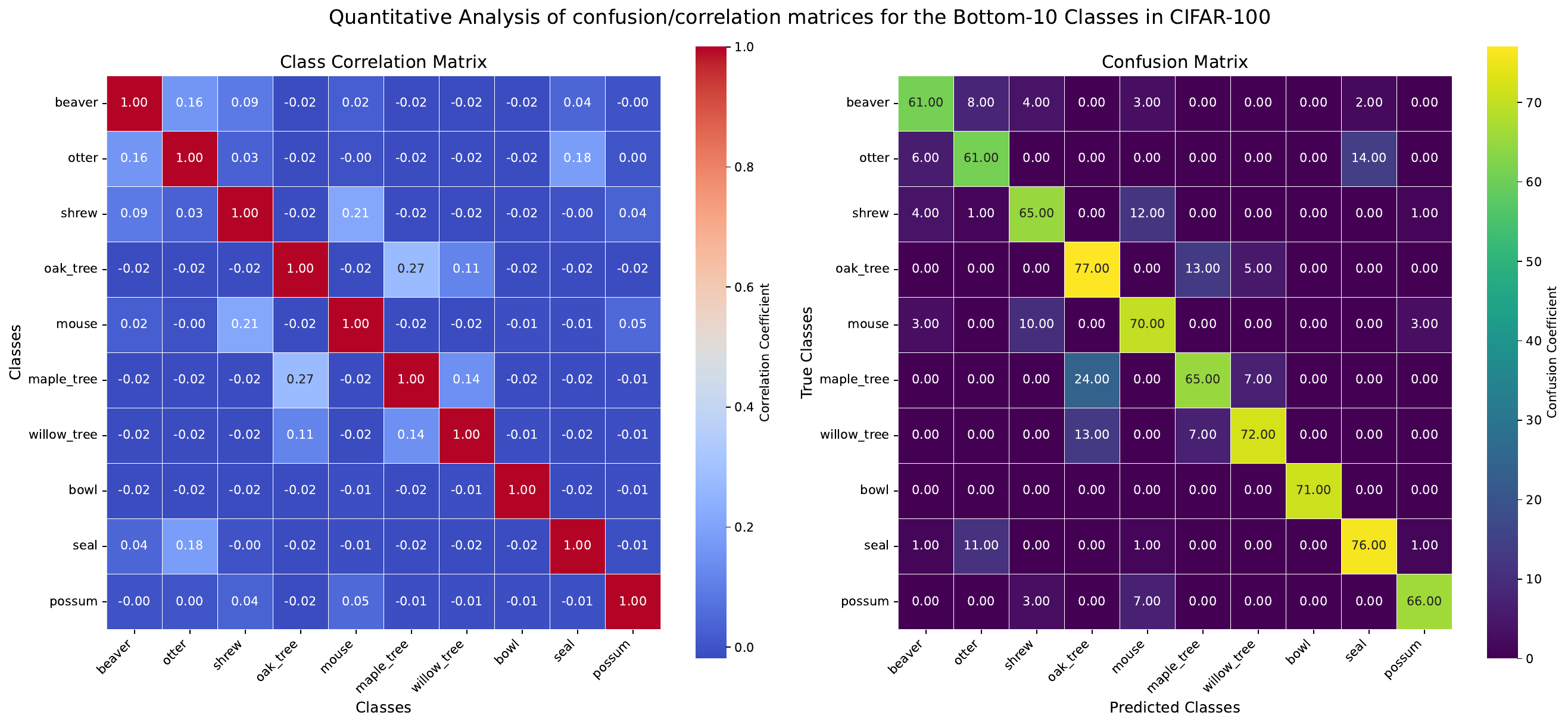}
    \includegraphics[height=0.25\paperheight]{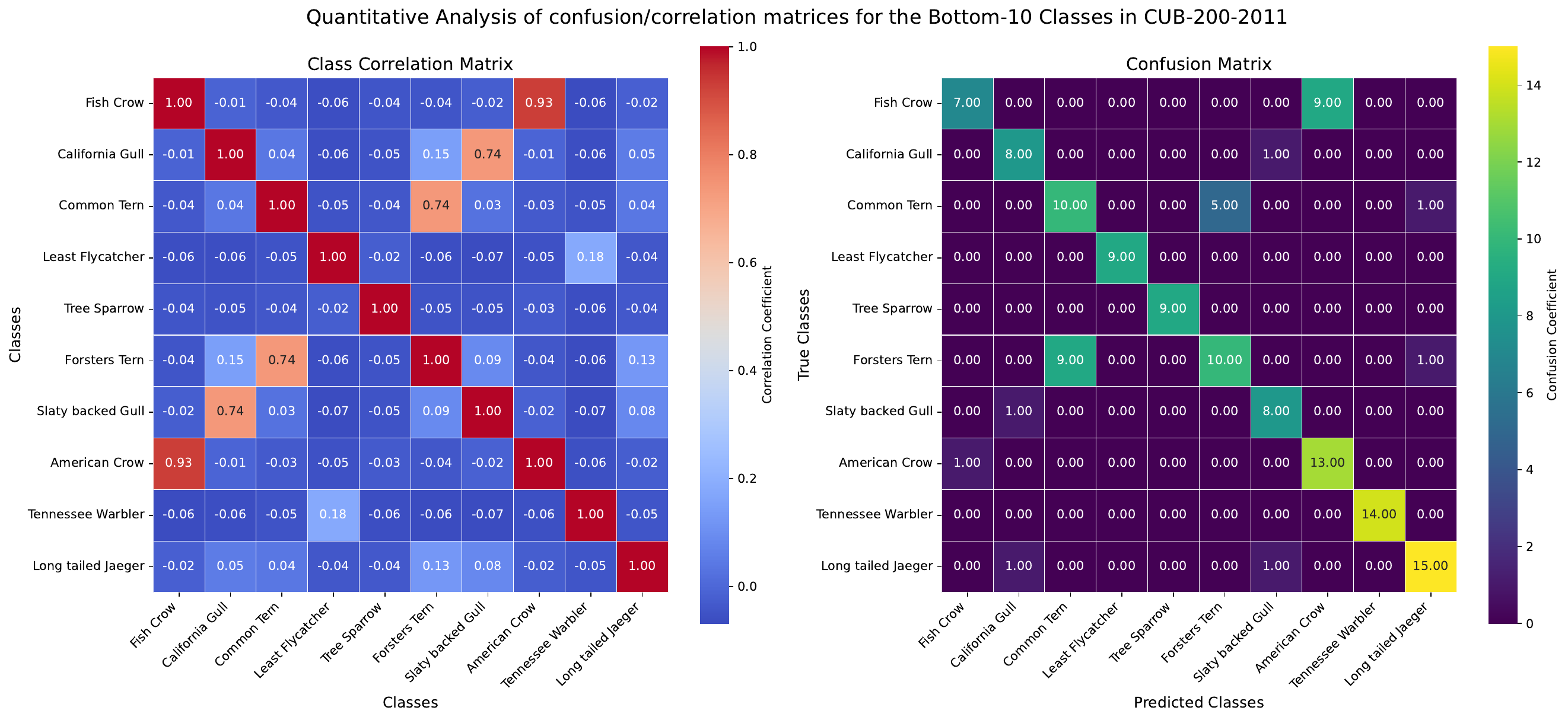}
    \includegraphics[height=0.25\paperheight]{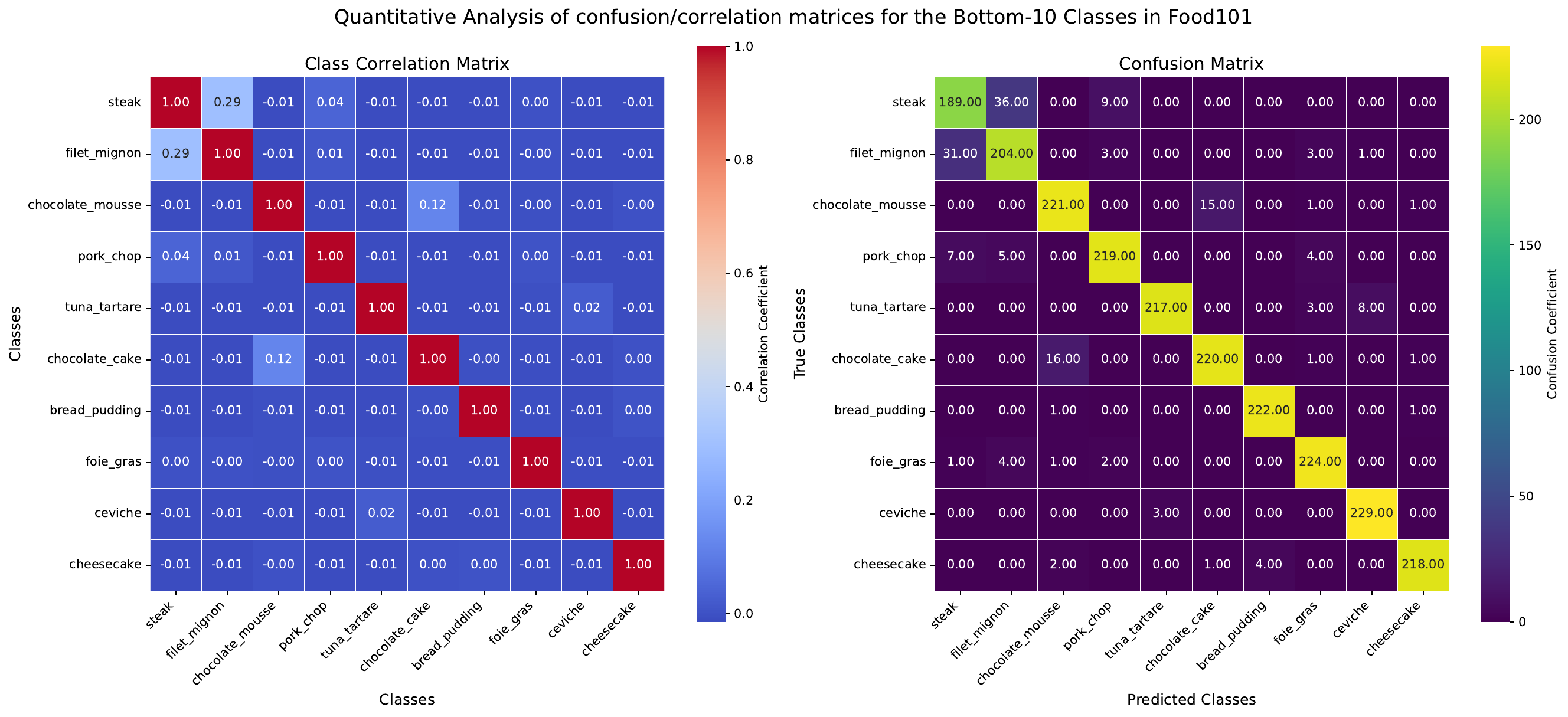}
    \caption{Qualitative analysis of \textit{Pearson's Correlation} disambiguation metric for the 10 most underperforming classes for CIFAR-100, CUB, and Food101. \name's heuristic does not require any human annotations, yet accurately approximates inter-class confusion. However, this heuristic is often over-sensitive to minute errors and is symmetric, leading to slightly suboptimal disambiguation.
    }
    \label{fig:visual_critic_ablations.2}
\end{figure*}

\begin{figure*}
    \centering
    \textbf{Qualitative analysis of `Top-k pseudo-confusion' as a disambiguation heuristic.}\par\medskip
    \includegraphics[height=0.25\paperheight]{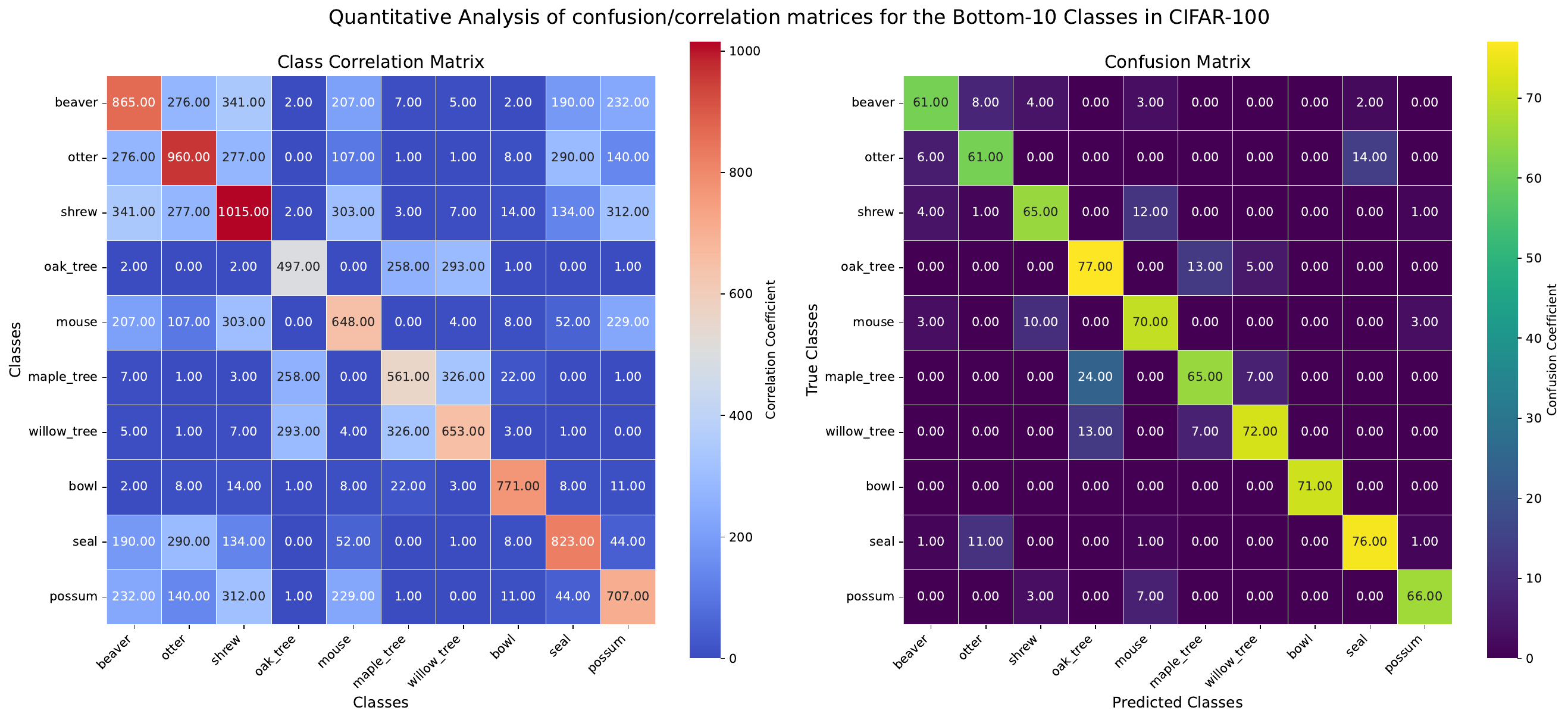}
    \includegraphics[height=0.25\paperheight]{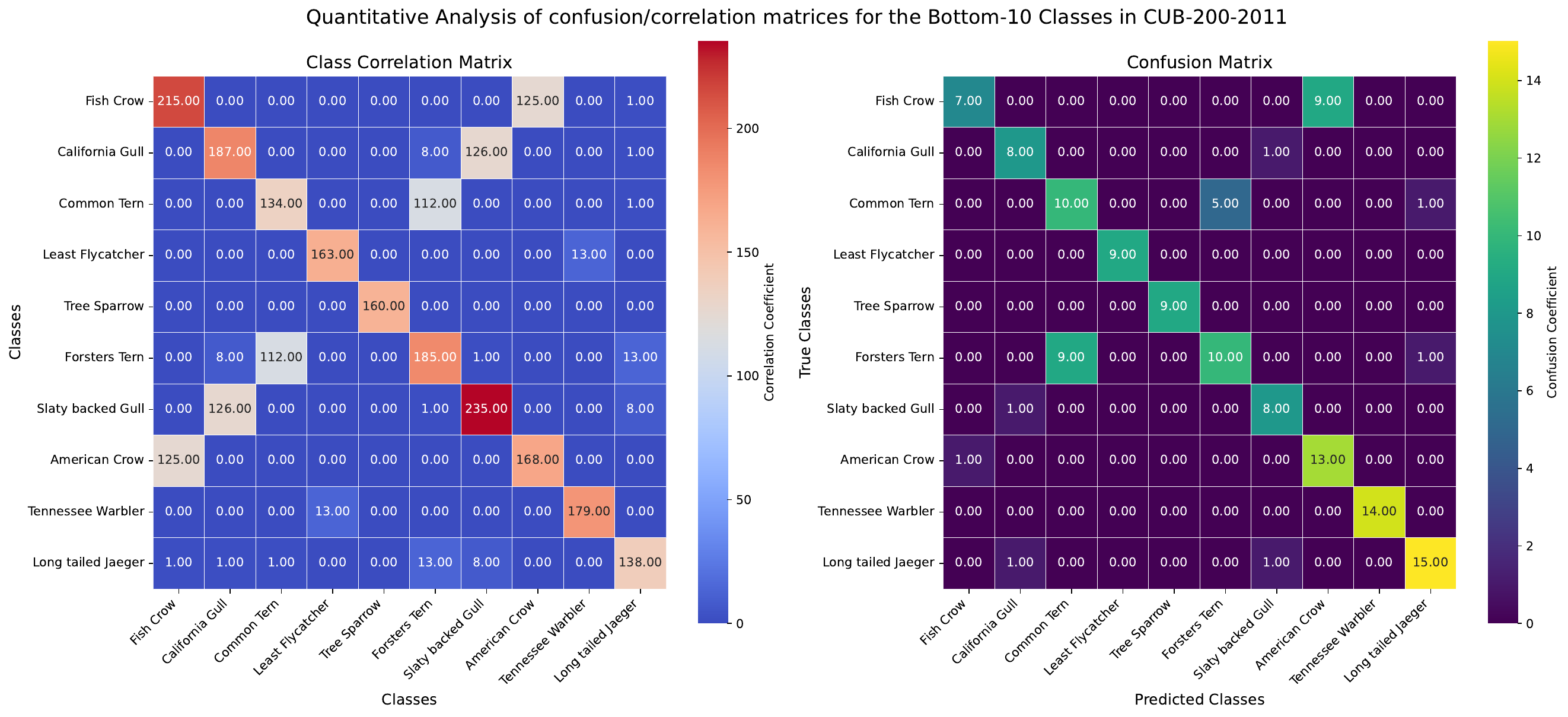}
    \includegraphics[height=0.25\paperheight]{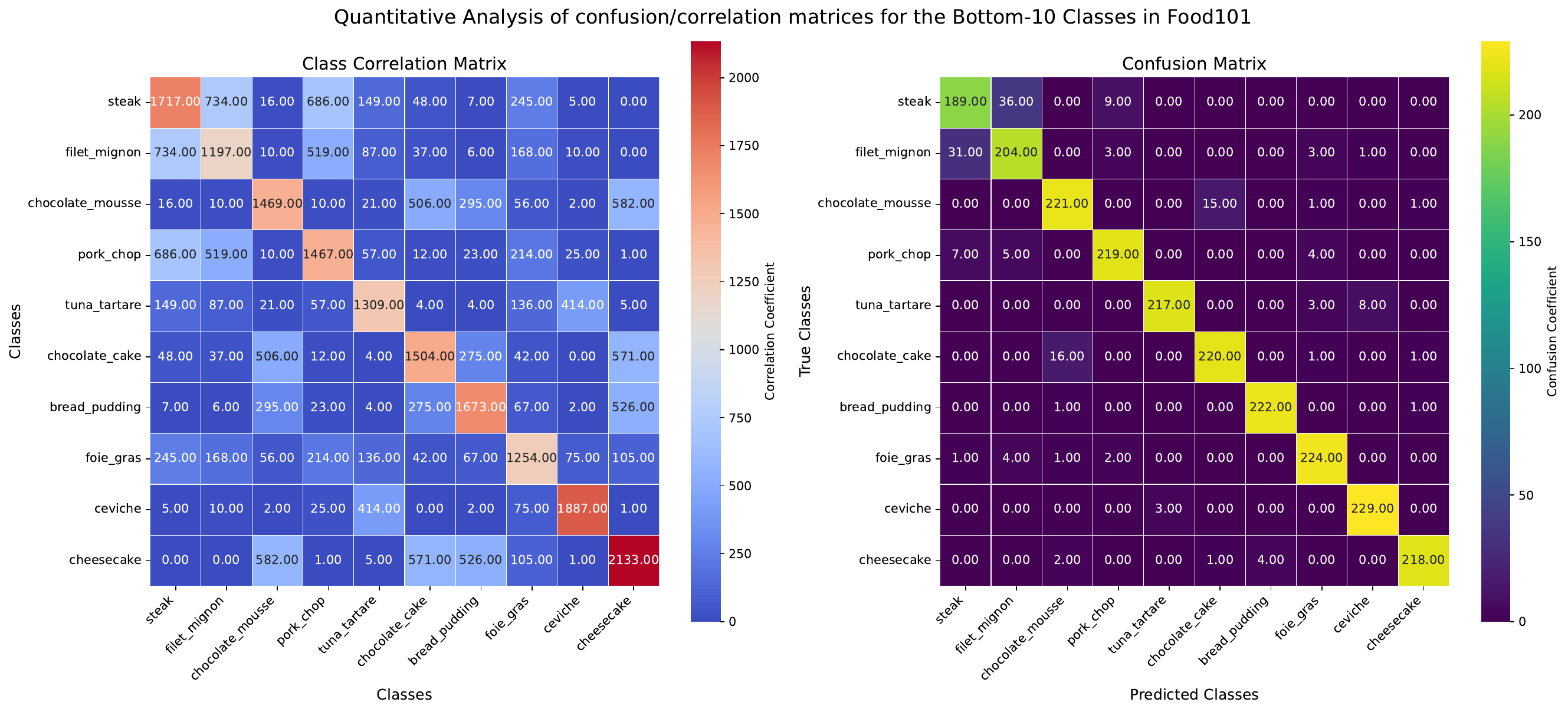}
    \caption{Qualitative analysis of the \textit{Top-k pseudo-confusion} disambiguation heuristic calculated for the 10 most underperforming classes for CIFAR-100, CUB, and Food101. After computing the top-k class scores $\text{topk}(\hat{\mathbf{y}})[:, :k]$, we compute the confusion matrix by incrementing the $(i, j)$ value if $y_i$ and $y_j$ class occur in the top-k entries for an image. Top-k pseudo-confusion tends to be sensitive to minute errors which leads to slight sub-optimality.
    }
    \label{fig:visual_critic_ablations.2.1}
\end{figure*}

\begin{figure*}
    \centering
    \includegraphics[height=\linewidth]{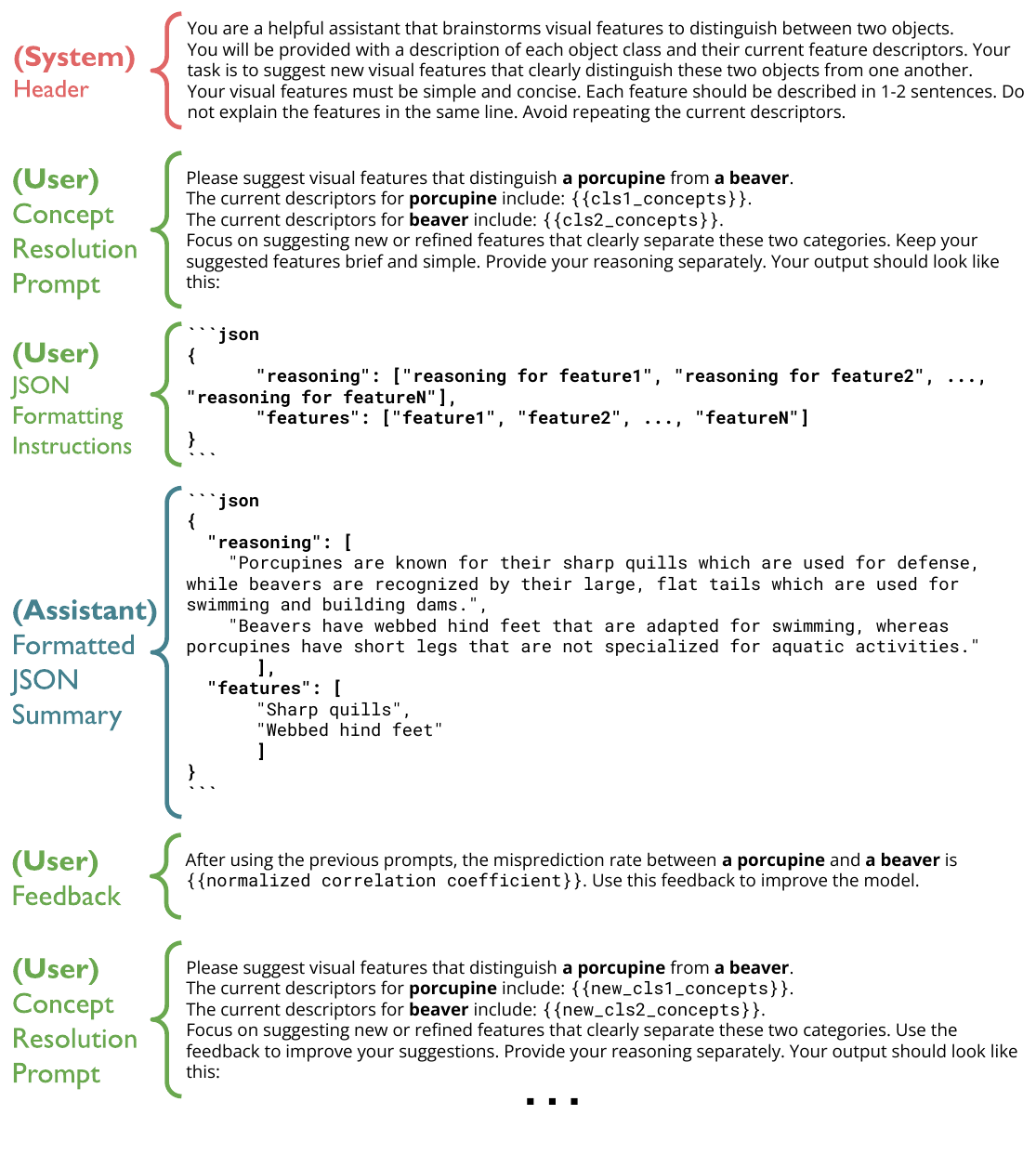}
    \caption{\name's prompt with history conditioning. We find that history conditioning is beneficial when the same disambiguation pair is repeatedly identified.}
    \label{fig:zero-shot-prompts}
\end{figure*}

\begin{sidewaysfigure*}
    \centering
    \includegraphics[width=\linewidth]{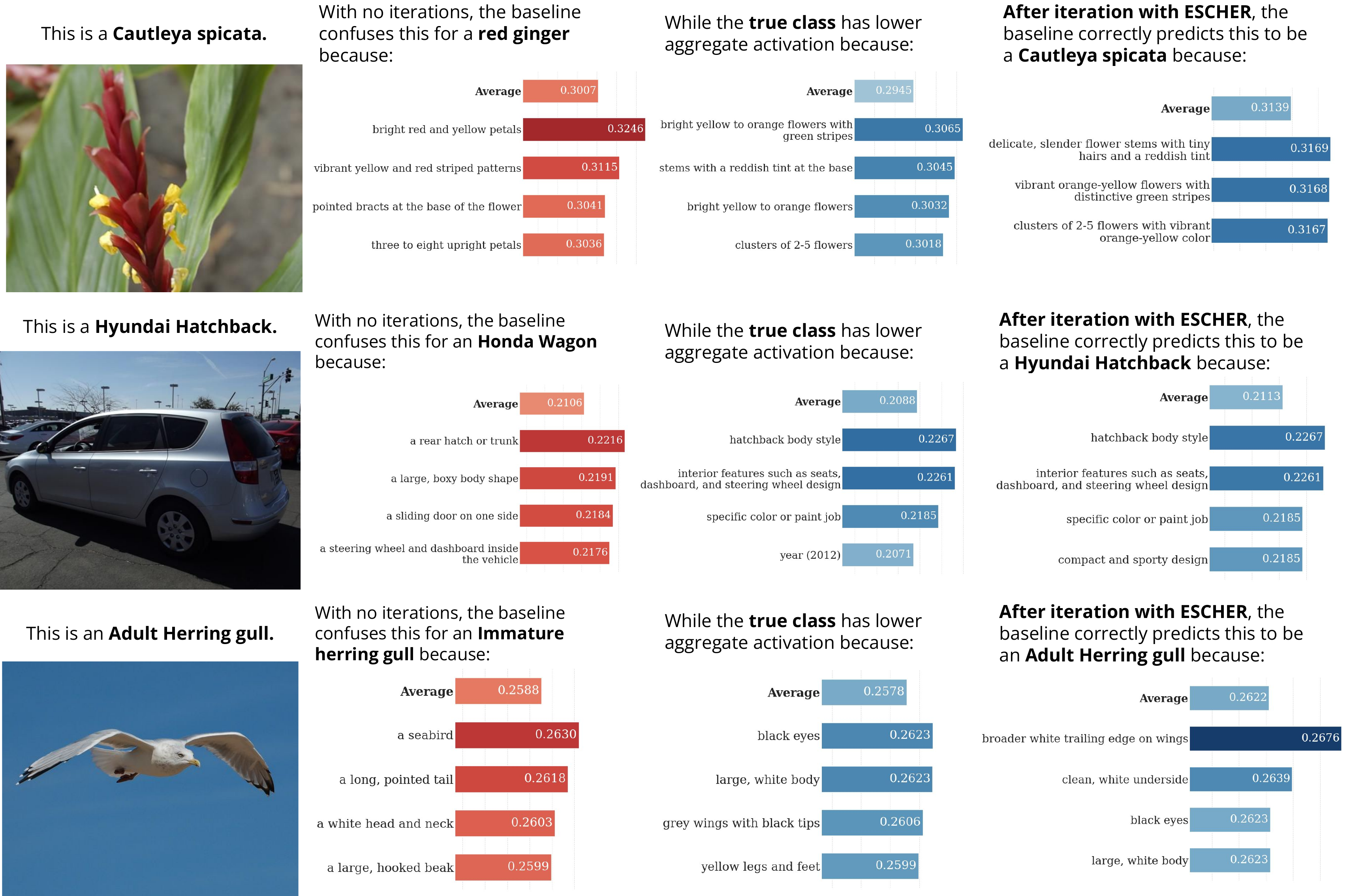}
    \caption{Qualitative analysis of the concept iteration progress on Oxford Flowers, Stanford Cars, and NABirds respectively for CbD+\name. We show samples of mispredicted classes for the zero-shot baseline (CbD) along with the confidence scores. With the initial concepts, the fine-grained class is misidentified. However, after iterating with \name, the baseline model is able to predict the correct class.}
    \label{fig:qualitative-results.1}
\end{sidewaysfigure*}

\subsection{Practical considerations for using \name}\label{sec:a.eschers-speed}

\paragraph{Experimental setup.} Evolving concepts with \name does not require a large computational footprint. We run our experiments on a server node with 10 NVIDIA A40 GPUs (each with 40 GB of VRAM), which allowed us to parallelize experiments. We were also able to replicate our experiments on a single NVIDIA RTX 2080 Ti GPU (12 GB of VRAM). For queries to large language models (LLMs), we primarily use externally hosted large language models (such as \texttt{gpt-3.5-turbo-0125}). For local models (e.g. \texttt{meta-llama/Llama-3.3-70B-Instruct}), we use \texttt{vLLM} \citep{kwon2023efficient} to host a local server. However, our framework is compatible with any LLM inference framework that allows hosting an OpenAI compliant server.

\paragraph{LLM queries per experiment.} For reference, each iteration dispatches about 100 queries (depending on the sub-sampling hyper-parameters). However, with history conditioning enabled, the size of each call to the LLM increases linearly with the number of iterations. Assuming each query is approximately 500 tokens, and each additional history entry adds 100 tokens, we can estimate the total token usage for 50 iterations to be around 128,000 tokens if we only consider just a single query per iteration and around 12.8M tokens for each experiment (approximately an hour).

\paragraph{Per iteration metrics.} Typically, a single \name iteration  can take 2 to 15 minutes for all CMBs on a single GPU. \name largely spends this time optimizing the CBM adapter. Another expensive operation -- generating disambiguation concepts with the LLM -- is parallelizable and finishes in 1-2 minutes. The number of concepts per class depends on the CBM and the dataset complexity. Each disambiguation query typically adds 2–5 concepts. We were able to comfortably fit an adapter with up to 5000 attributes (maximum we tested) on an NVIDIA RTX 2080 Ti GPU.

\paragraph{Choosing disambiguation heuristics.} In addition to the observations presented in \S \ref{sec:a.multiple_critics}, Figure \ref{fig:visual_critic_ablations.2}, and Figure \ref{fig:visual_critic_ablations.2.1} which highlight various advantages and shortcomings of each disambiguation heuristic, we find that, generally, grid-searching for the best disambiguation heuristic and hyper-parameters is surprisingly practical, when running for low number of iterations with a local language model (around 10 iterations).

\paragraph{Confusion matrix runtime.} The confusion score calculation is perfectly parallelizable, and is vectorized. Our naive numpy/scipy implementation takes roughly 6 seconds for our largest dataset (400 classes). Moreover, we can exploit properties of specific heuristics to derive further speedups (e.g. symmetric confusion matrix for \textit{Pearson's correlation}).

\subsection{\name's convergence properties}\label{sec:a.intuitive-liblearning-explanation}
Intuitively, \name employs the concept library to simultaneously bootstrap the CBM and LLM: A CBM with a specialized concept library produces more fine-grained class disambiguation feedback, prompting the LLM to uncover even finer concepts, which leads to an even more specialized library for the next iteration. Like other library learning algorithms \citep{Wong2021LeveragingLT, ellis2021dreamcoder, grand2024lilo}, ideally -- while the LLM disambiguates concepts well and the CBM remains sensitive to them -- this self-reinforcing loop continues until no further relevant concepts can be identified.  Empirical results suggest that \name discovers relevant concepts for many datasets. A deeper exploration of library learning's optimization properties are presented in \citep{ellis2021dreamcoder}.

\subsection{Bayesian formulation for CBMs}\label{sec:bayesian-formulation-cbm}
We can abstractly define the fitness of an adapter as the likelihood $p_{\Con}(\mathcal{D}|\Adapter)$ of the adapter generating the dataset $\mathcal{D}$. Not every concept will contribute to the final class assignment. Hence, we regularize the adapter by imposing a prior probability distribution $p_{\Con}(\Adapter)$ that is enforced by sampling concepts via an LLM, leveraging the prior knowledge of the LLM to filter out irrelevant concepts. Now, the problem of training a concept-bottleneck model can be expressed as a maximum a posteriori (MAP) estimation problem:
\begin{align}\label{eq:cbm-training}
\Adapter^\star &= \arg\max_{\Adapter} p_{\Con}(\Adapter|\mathcal{D}) \\\nonumber
&= \arg\max_{\Adapter} \underbrace{p_{\Con}(\mathcal{D}|\Adapter)}_{\text{optimize}} \cdot \underbrace{p_{\mathcal{C}}(\Adapter)}_{\text{regularizer}}
\end{align}

\subsection{History Conditioning}\label{sec:a.more_history}
Due to different training objectives and operational modalities, the concepts proposed by the LLM and the VLM's interpretation of concepts are bound to be misaligned frequently. In such cases, \name often needs to disambiguate the same pair of classes multiple times. Also, it is possible for each class disambiguation query to cause collisions with other classes in later iterations.  This motivates the need to keep track of past LLM proposals to each concept set as well as the VLM's response to each of the changes. \name implements this using the \textproc{InitializeHistory} and the \textproc{UpdateHistory} functions.

\noindent\textproc{InitializeHistory}. This function takes as input the number of classes $\mathcal{Y}$ and the number of iterations $T$ and constructs a data structure to hold the list of descriptors for a pair of classes $(i, j)$ at iteration $t$ as well as the updated class-confusion heuristic generated by the VLM at iteration $t+1$.

\noindent\textproc{UpdateHistory}. This function plays two roles. First, it stores the list of new descriptors for the $i^{\text{th}}$ and $j^{\text{th}}$ class for the $t+1^{\text{th}}$ iteration after the class confusion resolution. Second, in the subsequent iteration, the updated class confusion score  for the $(i, j)$ pair is stored to measure the effectiveness of the concepts proposed in the class confusion resolution step.

\subsection{Additional backbone ablation with ViT-B/32.}\label{sec:a.additional-ablation}

We report observations after evolving concepts with \name using LM4CV and CbD with a new backbone LLM (4bit quantized \texttt{Llama-3.3-70B-Instruct}) and VLM (ViT-B/32). The results are presented in Table~\ref{tab:r.clipb32}. Although overall accuracy is lower due to weaker backbone models, iterating with \name leads to better performance than relying on a fixed set of concepts.

\begin{table}[htbp]
\centering
\adjustbox{max width=0.45\textwidth}{
\begin{tabular}{lc|cc|cc}
\toprule
ViT-B/32 & CLIP & LM4CV & LM4CV & CbD & CbD \\
& & & +\name &  & +\name \\
\midrule
CIFAR-100       & 59.60 & 78.50 & \textbf{78.71} & 62.50 & \textbf{64.10} \\
CUB-200-2011    & 52.83 & 63.62 & \textbf{67.64} & 54.17 & \textbf{55.67} \\
Food101         & 77.23 & 87.95 & \textbf{88.09} & 79.99 & \textbf{80.36} \\
NABirds         & 39.69 & 57.99 & \textbf{59.24} & \textbf{41.48} & \textbf{41.48} \\
Stanford Cars   & 59.13 & 75.54 & \textbf{75.56} & 57.40 & \textbf{60.62} \\
\bottomrule
\end{tabular}
}
\caption{
Top-1 accuracy for evolving \name with a weaker LLM (Llama-3.3-70B-4bit) and visual critic (ViT-B/32). \name consistently improves the performance of LM4CV and CbD across datasets.
}
\label{tab:r.clipb32}
\end{table}

\clearpage
{
    \small
    \bibliographystyle{ieeenat_fullname}
    \bibliography{neurosymbolic}

\begin{thebibliography}{34}
\providecommand{\natexlab}[1]{#1}
\providecommand{\url}[1]{\texttt{#1}}
\expandafter\ifx\csname urlstyle\endcsname\relax
  \providecommand{\doi}[1]{doi: #1}\else
  \providecommand{\doi}{doi: \begingroup \urlstyle{rm}\Url}\fi

\bibitem[Austin et~al.(2021)Austin, Odena, Nye, Bosma, Michalewski, Dohan,
  Jiang, Cai, Terry, Le, and Sutton]{austin2021programsynthesislargelanguage}
Jacob Austin, Augustus Odena, Maxwell Nye, Maarten Bosma, Henryk Michalewski,
  David Dohan, Ellen Jiang, Carrie Cai, Michael Terry, Quoc Le, and Charles
  Sutton.
\newblock Program synthesis with large language models, 2021.

\bibitem[Beyer et~al.(2024)Beyer, Steiner, Pinto, Kolesnikov, Wang, Salz,
  Neumann, Alabdulmohsin, Tschannen, Bugliarello, et~al.]{beyer2024paligemma}
Lucas Beyer, Andreas Steiner, Andr{\'e}~Susano Pinto, Alexander Kolesnikov,
  Xiao Wang, Daniel Salz, Maxim Neumann, Ibrahim Alabdulmohsin, Michael
  Tschannen, Emanuele Bugliarello, et~al.
\newblock Paligemma: A versatile 3b vlm for transfer.
\newblock \emph{arXiv preprint arXiv:2407.07726}, 2024.

\bibitem[Bossard et~al.(2014)Bossard, Guillaumin, and
  Van~Gool]{bossard2014food}
Lukas Bossard, Matthieu Guillaumin, and Luc Van~Gool.
\newblock Food-101--mining discriminative components with random forests.
\newblock In \emph{Computer vision--ECCV 2014: 13th European conference,
  zurich, Switzerland, September 6-12, 2014, proceedings, part VI 13}, pages
  446--461. Springer, 2014.

\bibitem[Brown et~al.(2020)Brown, Mann, Ryder, Subbiah, Kaplan, Dhariwal,
  Neelakantan, Shyam, Sastry, Askell, et~al.]{brown2020language}
Tom Brown, Benjamin Mann, Nick Ryder, Melanie Subbiah, Jared~D Kaplan, Prafulla
  Dhariwal, Arvind Neelakantan, Pranav Shyam, Girish Sastry, Amanda Askell,
  et~al.
\newblock Language models are few-shot learners.
\newblock \emph{Advances in neural information processing systems},
  33:\penalty0 1877--1901, 2020.

\bibitem[Chiquier et~al.(2024)Chiquier, Mall, and
  Vondrick]{chiquier2024evolving}
Mia Chiquier, Utkarsh Mall, and Carl Vondrick.
\newblock Evolving interpretable visual classifiers with large language models.
\newblock \emph{arXiv preprint arXiv:2404.09941}, 2024.

\bibitem[Ellis et~al.(2021)Ellis, Wong, Nye, Sabl{\'e}-Meyer, Morales, Hewitt,
  Cary, Solar-Lezama, and Tenenbaum]{ellis2021dreamcoder}
Kevin Ellis, Catherine Wong, Maxwell Nye, Mathias Sabl{\'e}-Meyer, Lucas
  Morales, Luke Hewitt, Luc Cary, Armando Solar-Lezama, and Joshua~B Tenenbaum.
\newblock Dreamcoder: Bootstrapping inductive program synthesis with wake-sleep
  library learning.
\newblock In \emph{Proceedings of the 42nd acm sigplan international conference
  on programming language design and implementation}, pages 835--850, 2021.

\bibitem[Grand et~al.(2024)Grand, Wong, Bowers, Olausson, Liu, Tenenbaum, and
  Andreas]{grand2024lilo}
Gabriel Grand, Lionel Wong, Maddy Bowers, Theo~X. Olausson, Muxin Liu,
  Joshua~B. Tenenbaum, and Jacob Andreas.
\newblock {LILO}: Learning interpretable libraries by compressing and
  documenting code.
\newblock In \emph{The Twelfth International Conference on Learning
  Representations}, 2024.

\bibitem[Grattafiori et~al.(2024)Grattafiori, Dubey, Jauhri, Pandey, Kadian,
  Al-Dahle, Letman, Mathur, Schelten, Vaughan, et~al.]{grattafiori2024llama}
Aaron Grattafiori, Abhimanyu Dubey, Abhinav Jauhri, Abhinav Pandey, Abhishek
  Kadian, Ahmad Al-Dahle, Aiesha Letman, Akhil Mathur, Alan Schelten, Alex
  Vaughan, et~al.
\newblock The llama 3 herd of models.
\newblock \emph{arXiv preprint arXiv:2407.21783}, 2024.

\bibitem[Grayeli et~al.(2024)Grayeli, Sehgal, Costilla-Reyes, Cranmer, and
  Chaudhuri]{grayeli2024symbolic}
Arya Grayeli, Atharva Sehgal, Omar Costilla-Reyes, Miles Cranmer, and Swarat
  Chaudhuri.
\newblock Symbolic regression with a learned concept library.
\newblock In \emph{Neural Information Processing Systems (NeurIPS)}, 2024.

\bibitem[Hu et~al.(2024)Hu, Iscen, Jain, Kipf, Yue, Ross, Schmid, and
  Fathi]{hu2024scenecraft}
Ziniu Hu, Ahmet Iscen, Aashi Jain, Thomas Kipf, Yisong Yue, David~A Ross,
  Cordelia Schmid, and Alireza Fathi.
\newblock Scenecraft: An llm agent for synthesizing 3d scenes as blender code.
\newblock In \emph{Forty-first International Conference on Machine Learning},
  2024.

\bibitem[Jia et~al.(2021)Jia, Yang, Xia, Chen, Parekh, Pham, Le, Sung, Li, and
  Duerig]{jia2021scaling}
Chao Jia, Yinfei Yang, Ye Xia, Yi-Ting Chen, Zarana Parekh, Hieu Pham, Quoc Le,
  Yun-Hsuan Sung, Zhen Li, and Tom Duerig.
\newblock Scaling up visual and vision-language representation learning with
  noisy text supervision.
\newblock In \emph{International conference on machine learning}, pages
  4904--4916. PMLR, 2021.

\bibitem[Kim et~al.(2021)Kim, Kim, Seo, and Yoon]{kim2021xprotonet}
Eunji Kim, Siwon Kim, Minji Seo, and Sungroh Yoon.
\newblock Xprotonet: diagnosis in chest radiography with global and local
  explanations.
\newblock In \emph{Proceedings of the IEEE/CVF conference on computer vision
  and pattern recognition}, pages 15719--15728, 2021.

\bibitem[Koh et~al.(2020)Koh, Nguyen, Tang, Mussmann, Pierson, Kim, and
  Liang]{koh2020conceptbottleneckmodels}
Pang~Wei Koh, Thao Nguyen, Yew~Siang Tang, Stephen Mussmann, Emma Pierson, Been
  Kim, and Percy Liang.
\newblock Concept bottleneck models, 2020.

\bibitem[Krause et~al.(2013)Krause, Stark, Deng, and Fei-Fei]{krause20133d}
Jonathan Krause, Michael Stark, Jia Deng, and Li Fei-Fei.
\newblock 3d object representations for fine-grained categorization.
\newblock In \emph{Proceedings of the IEEE international conference on computer
  vision workshops}, pages 554--561, 2013.

\bibitem[Krizhevsky et~al.(2012)Krizhevsky, Sutskever, and
  Hinton]{krizhevsky2009learning}
Alex Krizhevsky, Ilya Sutskever, and Geoffrey~E Hinton.
\newblock Imagenet classification with deep convolutional neural networks.
\newblock In \emph{Advances in Neural Information Processing Systems}. Curran
  Associates, Inc., 2012.

\bibitem[Kwon et~al.(2023)Kwon, Li, Zhuang, Sheng, Zheng, Yu, Gonzalez, Zhang,
  and Stoica]{kwon2023efficient}
Woosuk Kwon, Zhuohan Li, Siyuan Zhuang, Ying Sheng, Lianmin Zheng, Cody~Hao Yu,
  Joseph~E. Gonzalez, Hao Zhang, and Ion Stoica.
\newblock Efficient memory management for large language model serving with
  pagedattention.
\newblock In \emph{Proceedings of the ACM SIGOPS 29th Symposium on Operating
  Systems Principles}, 2023.

\bibitem[Lake et~al.(2015)Lake, Salakhutdinov, and Tenenbaum]{lake2015human}
Brenden~M Lake, Ruslan Salakhutdinov, and Joshua~B Tenenbaum.
\newblock Human-level concept learning through probabilistic program induction.
\newblock \emph{Science}, 350\penalty0 (6266):\penalty0 1332--1338, 2015.

\bibitem[Li et~al.(2018)Li, Liu, Chen, and Rudin]{li2018deep}
Oscar Li, Hao Liu, Chaofan Chen, and Cynthia Rudin.
\newblock Deep learning for case-based reasoning through prototypes: A neural
  network that explains its predictions.
\newblock In \emph{Proceedings of the AAAI Conference on Artificial
  Intelligence}, 2018.

\bibitem[Liu et~al.(2024)Liu, Li, Wu, and Lee]{liu2024visual}
Haotian Liu, Chunyuan Li, Qingyang Wu, and Yong~Jae Lee.
\newblock Visual instruction tuning.
\newblock \emph{Advances in neural information processing systems}, 36, 2024.

\bibitem[Menon and Vondrick(2023)]{DBLP:conf/iclr/MenonV23}
Sachit Menon and Carl Vondrick.
\newblock Visual classification via description from large language models.
\newblock In \emph{The Eleventh International Conference on Learning
  Representations, {ICLR} 2023, Kigali, Rwanda, May 1-5, 2023}. OpenReview.net,
  2023.

\bibitem[Nilsback and Zisserman(2008)]{nilsback2008automated}
Maria-Elena Nilsback and Andrew Zisserman.
\newblock Automated flower classification over a large number of classes.
\newblock In \emph{2008 Sixth Indian conference on computer vision, graphics \&
  image processing}, pages 722--729. IEEE, 2008.

\bibitem[Nye et~al.(2021)Nye, Andreassen, Gur-Ari, Michalewski, Austin, Bieber,
  Dohan, Lewkowycz, Bosma, Luan, Sutton, and
  Odena]{nye2021workscratchpadsintermediatecomputation}
Maxwell Nye, Anders~Johan Andreassen, Guy Gur-Ari, Henryk Michalewski, Jacob
  Austin, David Bieber, David Dohan, Aitor Lewkowycz, Maarten Bosma, David
  Luan, Charles Sutton, and Augustus Odena.
\newblock Show your work: Scratchpads for intermediate computation with
  language models, 2021.

\bibitem[Pratt et~al.(2022)Pratt, Liu, and Farhadi]{Pratt2022WhatDA}
Sarah Pratt, Rosanne Liu, and Ali Farhadi.
\newblock What does a platypus look like? generating customized prompts for
  zero-shot image classification.
\newblock \emph{2023 IEEE/CVF International Conference on Computer Vision
  (ICCV)}, pages 15645--15655, 2022.

\bibitem[Radford et~al.(2021)Radford, Kim, Hallacy, Ramesh, Goh, Agarwal,
  Sastry, Askell, Mishkin, Clark, et~al.]{radford2021learning}
Alec Radford, Jong~Wook Kim, Chris Hallacy, Aditya Ramesh, Gabriel Goh,
  Sandhini Agarwal, Girish Sastry, Amanda Askell, Pamela Mishkin, Jack Clark,
  et~al.
\newblock Learning transferable visual models from natural language
  supervision.
\newblock In \emph{International conference on machine learning}, pages
  8748--8763. PMLR, 2021.

\bibitem[Saha et~al.(2024)Saha, Van~Horn, and Maji]{saha2024improved}
Oindrila Saha, Grant Van~Horn, and Subhransu Maji.
\newblock Improved zero-shot classification by adapting vlms with text
  descriptions.
\newblock In \emph{Proceedings of the IEEE/CVF Conference on Computer Vision
  and Pattern Recognition}, pages 17542--17552, 2024.

\bibitem[Schick et~al.(2023)Schick, Dwivedi-Yu, Dessì, Raileanu, Lomeli,
  Zettlemoyer, Cancedda, and Scialom]{schick2023toolformerlanguagemodelsteach}
Timo Schick, Jane Dwivedi-Yu, Roberto Dessì, Roberta Raileanu, Maria Lomeli,
  Luke Zettlemoyer, Nicola Cancedda, and Thomas Scialom.
\newblock Toolformer: Language models can teach themselves to use tools, 2023.

\bibitem[Tong et~al.(2024)Tong, Liu, Zhai, Ma, LeCun, and
  Xie]{tong2024eyeswideshutexploring}
Shengbang Tong, Zhuang Liu, Yuexiang Zhai, Yi Ma, Yann LeCun, and Saining Xie.
\newblock Eyes wide shut? exploring the visual shortcomings of multimodal llms,
  2024.

\bibitem[Van~Horn et~al.(2015)Van~Horn, Branson, Farrell, Haber, Barry,
  Ipeirotis, Perona, and Belongie]{van2015building}
Grant Van~Horn, Steve Branson, Ryan Farrell, Scott Haber, Jessie Barry, Panos
  Ipeirotis, Pietro Perona, and Serge Belongie.
\newblock Building a bird recognition app and large scale dataset with citizen
  scientists: The fine print in fine-grained dataset collection.
\newblock In \emph{Proceedings of the IEEE conference on computer vision and
  pattern recognition}, pages 595--604, 2015.

\bibitem[Wah et~al.(2023)Wah, Branson, Welinder, Perona, and
  Belongie]{wah2011caltech}
Catherine Wah, Steve Branson, Peter Welinder, Pietro Perona, and Serge
  Belongie.
\newblock The caltech-ucsd birds-200-2011 dataset, 2023.

\bibitem[Wong et~al.(2021)Wong, Ellis, Tenenbaum, and
  Andreas]{Wong2021LeveragingLT}
Catherine Wong, Kevin Ellis, Joshua~B. Tenenbaum, and Jacob Andreas.
\newblock Leveraging language to learn program abstractions and search
  heuristics.
\newblock In \emph{International Conference on Machine Learning}, 2021.

\bibitem[Yan et~al.(2023)Yan, Wang, Zhong, Dong, He, Lu, Wang, Shang, and
  McAuley]{yan2023learning}
An Yan, Yu Wang, Yiwu Zhong, Chengyu Dong, Zexue He, Yujie Lu, William~Yang
  Wang, Jingbo Shang, and Julian McAuley.
\newblock Learning concise and descriptive attributes for visual recognition.
\newblock In \emph{Proceedings of the IEEE/CVF International Conference on
  Computer Vision}, pages 3090--3100, 2023.

\bibitem[Yang et~al.(2023)Yang, Panagopoulou, Zhou, Jin, Callison-Burch, and
  Yatskar]{yang2023languagebottlelanguagemodel}
Yue Yang, Artemis Panagopoulou, Shenghao Zhou, Daniel Jin, Chris
  Callison-Burch, and Mark Yatskar.
\newblock Language in a bottle: Language model guided concept bottlenecks for
  interpretable image classification, 2023.

\bibitem[Yu et~al.(2022)Yu, Wang, Vasudevan, Yeung, Seyedhosseini, and
  Wu]{yu2022coca}
Jiahui Yu, Zirui Wang, Vijay Vasudevan, Legg Yeung, Mojtaba Seyedhosseini, and
  Yonghui Wu.
\newblock Coca: Contrastive captioners are image-text foundation models.
\newblock \emph{arXiv preprint arXiv:2205.01917}, 2022.

\bibitem[Zhang et~al.(2023)Zhang, Hu, Li, Huang, Deng, Li, Qiao, and
  Gao]{zhang2023prompt}
Renrui Zhang, Xiangfei Hu, Bohao Li, Siyuan Huang, Hanqiu Deng, Hongsheng Li,
  Yu Qiao, and Peng Gao.
\newblock Prompt, generate, then cache: Cascade of foundation models makes
  strong few-shot learners.
\newblock \emph{arXiv preprint arXiv:2303.02151}, 2023.

\end{thebibliography}
}

\end{document}